\documentclass{IEEEtran}
\usepackage{amsmath,amsfonts}
\usepackage{algorithmic}
\usepackage{algorithm}
\usepackage{array}
\usepackage{hyperref} 
\usepackage[caption=false,font=normalsize,labelfont=sf,textfont=sf]{subfig}
\usepackage{textcomp}
\usepackage{stfloats}
\usepackage{url}
\usepackage{verbatim}
\usepackage{graphicx}
\usepackage{cite}
\usepackage{amsmath,amssymb,amsfonts}
\usepackage{xcolor}
\usepackage{multirow}
\usepackage{array}
\usepackage{siunitx} % for \degree
\usepackage{gensymb} % for \degree
\usepackage{caption}
\usepackage{amssymb}
\usepackage{booktabs}
\usepackage{pifont}
\usepackage[table]{xcolor}
\begin{document}

\title{Simulating Sinogram-Domain Motion and Correcting Image-Domain Artifacts Using Deep Learning in HR-pQCT Bone Imaging}

\author{Farhan Sadik,~\IEEEmembership{Member,~IEEE}, Christopher L. Newman, Stuart J. Warden, and Rachel K. Surowiec
        % <-this % stops a space
\thanks{This work is supported by
National Institutes of Health (NIH/NIAMS P30 AR072581), and (LRP 1L30DK130133-0). This work involved human subjects in its research. Approval of all ethical and experimental procedures and protocols was granted by the Institutional Review Board of Indiana University (IRB protocol \#1707550885) and performed in line with the Declaration of Helsinki.}
\thanks{Farhan Sadik is with the Weldon School of Biomedical Engineering, Purdue University, West Lafayette, IN, USA. (e-mail: fsadik@purdue.edu)}
\thanks{Christopher L. Newman is with the Department of Radiology and Imaging Sciences, Indiana University School of Medicine, Indianapolis, IN, USA. (e-mail: chrnewma@iu.edu)}
\thanks{Stuart J. Warden is with the Department of Physical Therapy, School of Health \& Human Sciences, Indiana University Indianapolis, Indianapolis, IN, USA. (e-mail: stwarden@iu.edu)}
\thanks{Rachel Surowiec is with the Weldon School of Biomedical Engineering, Purdue University, West Lafayette, IN, USA. (e-mail: rsurowie@purdue.edu)}}

\markboth{Journal of \LaTeX\ Class Files,~Vol.~X, No.~X, August~2025}%
{Shell \MakeLowercase{\textit{Sadik et al.}}: A Sample Article Using IEEEtran.cls for IEEE Journals}


\maketitle

\begin{abstract}
Rigid-motion artifacts, such as cortical bone streaking and trabecular smearing, hinder in vivo assessment of bone microstructures in high-resolution peripheral quantitative computed tomography (HR-pQCT). Despite various motion grading techniques, no motion correction methods exist due to the lack of standardized degradation models.
We \textcolor{black}{optimize a conventional sinogram-based method to simulate motion artifacts in HR-pQCT images, creating paired datasets of motion-corrupted images and their corresponding ground truth, which enables seamless integration into supervised learning frameworks for motion correction.} As such, we propose an Edge-enhanced Self-attention Wasserstein Generative Adversarial Network with Gradient Penalty (ESWGAN-GP) to address motion artifacts in both simulated (source) and real-world (target) datasets. The model incorporates edge-enhancing skip connections to preserve trabecular edges and self-attention mechanisms to capture long-range dependencies, facilitating motion correction. A visual geometry group (VGG)-based perceptual loss is used to reconstruct fine micro-structural features. The ESWGAN-GP achieves a mean signal-to-noise ratio (SNR) of 26.78, structural similarity index measure (SSIM) of 0.81, and visual information fidelity (VIF) of 0.76 for the source dataset, while showing improved performance on the target dataset with an SNR of 29.31, SSIM of 0.87, and VIF of 0.81. \textcolor{black}{The proposed methods address a simplified representation of real-world motion that may not fully capture the complexity of in vivo motion artifacts. Nevertheless, because motion artifacts present one of the foremost challenges to more widespread adoption of this modality, these methods represent an important initial step toward implementing deep learning-based motion correction in HR-pQCT.}

%\textcolor{black}{Although the proposed method addresses a simplified representation of real-world motion and may not fully capture the complexity of in vivo motion artifacts, and is further constrained by architectural limitations, it nonetheless represents an important initial step toward deep learning-based motion correction in HR-pQCT}, potentially reducing patient rescans by up to 10\% per week, a major challenge for the broader adoption of this modality.

\end{abstract}


\section{Introduction}
\label{sec1}
\IEEEPARstart{T}{o} date, X-ray-based modalities remain the gold standard for non-invasive bone quality measurement and continue to be widely used for the detection of bone fragility and fractures \cite{Seeman2008}. However, diseases such as osteoporosis, type 2 diabetes, chronic kidney disease (CKD), and rheumatoid arthritis \cite{Surowiec2022} alter bone matrix mineralization, resulting in distinct changes to cortical and trabecular microarchitecture. These changes necessitate the use of high-resolution imaging techniques, such as micro-CT \cite{Pereira2016}, peripheral quantitative computed tomography (pQCT) \cite{Stagi2016}, and high-resolution \textcolor{black}{pQCT} (HR-pQCT), for detailed assessment. HR-pQCT is a 
form of 3D longitudinal cone beam CT (CBCT) imaging that primarily focuses on peripheral skeletal sites such as the 
distal radius and tibia \cite{Gazzotti2023}. It can analyze high-resolution ($\approx$61–82 $\mu$m isotropic voxel size) micro-architecture of cortical bone, the dense outer layer of long bones, and trabecular bone, the lattice-like structure within the marrow cavity. \textcolor{black}{The imaging is performed} with nominal radiation ($\sim$3 $\mu$Sv) \cite{Lee2014}, making it 
attractive for the pediatric population, patients who need to minimize radiation, and for use in clinical trials where imaging multiple time points is desired.

However, the relatively long acquisition time ($\approx$2 minutes depending on scanner generation and settings) combined with the super-resolution nature of HR-pQCT makes it sensitive to motion artifacts where 
a small amount of motion can diminish the micro-structure information. Current protocol against subject-specific motion
generally includes re-scanning of patients, which may not be feasible in ill patients or patients suffering from tremors, twitches, and spasms \cite{Sode2011}, and may not be permissible in a busy clinical workflow. Furthermore, users of this technology have identified retrospective motion correction 
as a critical need requiring a solution \cite{Gabel2023}.

Motion artifacts, estimated to affect up to 23\% of first-generation HR-pQCT scans \cite{Pialat2012}, are traditionally identified through subjective grading using reference images provided by the manufacturer. This involves scoring the image based on the extent of motion corruption, with a score of 1 indicating the highest image quality and a score of 5 representing the lowest. Specifically, the grading reflects the progression of motion artifacts, ranging from minor horizontal/vertical streaks (score 1) to major streaks (score 5), with disruptions in cortical continuity and trabecular smearing seen in score 4 and up \cite{Pialat2012}. Sode et al. \cite{Sode2011} were the first to propose a quantitative metric for measuring motion artifacts using raw sinogram data. They used image similarity metrics, including the sum of squared differences (SSD) and normalized cross-correlation (NCC), between two aligned projections at 0\degree{} and 180\degree{}. The underlying assumption was that motion significantly alters the acquisition mode, magnitude, and timing. Ideally, projections at 0\degree{} should match those at 180\degree{}, and any discrepancies would indicate motion artifacts. More recently, deep learning-based methods have emerged as effective tools for motion grading in HR-pQCT. Both Walle et al. \cite{Walle2023}, and Benedikt et al. \cite{Benedikt2024} introduced convolutional neural networks (CNNs) to identify five distinct levels of motion grades, addressing limitations of subjective manual grading and improving study comparability. Even so, subject-specific grading remains the prevalent method among users, and there are currently no available reproducible codes for deep learning-based automated grading systems.

\textcolor{black}{An early effort toward motion correction in HR-pQCT, aimed at simulating motion-induced artifacts, was presented by Pauchard et al. \cite{5193053} in the context of first-generation scanners (with a resolution of approximately 82 $\mu$m). In this work, a modified form of the Helgason-Ludwig consistency conditions (HLCC) was employed to independently simulate in-plane translational and rotational motion. Specifically, translational motion was inferred from the first moment, while rotational motion was identified through the second moment of the HLCC. Subsequently, the authors evaluated the impact of these motion-induced errors on bone microarchitecture quantification tasks, such as the measurement of cortical thickness, trabecular number, and related parameters. Building upon their earlier work, Pauchard et al. \cite{Pauchard2011} conducted a subsequent study that incorporated longitudinal translation alongside in-plane rotation and translation. To physically replicate these motion parameters, the authors employed a custom-designed mechanical apparatus applied to cadaveric bone specimens. These approaches laid the groundwork for systematically embedding simulated motion artifacts into ground truth datasets, thereby facilitating the creation of large, annotated datasets essential for training deep learning models focused on motion correction.}
Nonetheless, only a single abstract on motion correction has been identified in HR-pQCT literature, 
which employed a Cycle-consistent Generative Adversarial Network (Cycle GAN) to learn the mapping from low-quality blurred to 
deblurred high-quality images \cite{steiner2022correction}. However, this motion deblurring model did not account for cortical breaks, which constitute the majority, if not the entirety, of motion-related distortions. Furthermore, motion in HR-pQCT is primarily caused by patient movement\textcolor{black}{\cite{5193053, Pauchard2011}} rather than sensor displacement; thus, a deblurring model may not be the most suitable approach for accurately capturing motion artifacts. 

The simulation of patient motion and the development of correction models have been extensively investigated in X-ray and conventional CT imaging literature. Accurate modeling of patient motion is an essential step before correction, as it spares researchers the painstaking task of collecting large datasets that include both motion-corrupted and motion-free images for individual patients. Some models are based on motion dynamics such as rotation, translation, and oscillation of the imaging object \cite{Su2021}, while others rely on prospectively collected six-degree-of-freedom motion data using optical tracking systems \cite{Kim2015}, or account for changes in the attenuation field \cite{Boigne2022}. Nevertheless, none of these methods have been implemented for HR-pQCT. 

A comparable situation can be observed in Coronary CT Angiography, where a step-and-shoot acquisition method is employed, similar to HR-pQCT. The CoMoFACT framework, proposed by Lossau et al. \cite{LossauneElss2019, Lossau2019}, provides a methodology for simulating motion artifacts in originally motion-free images, generating corresponding 2D image pairs in Cardiac CT. A CNN is then employed to estimate the motion vector field, which is integrated into an iterative motion compensation algorithm. Maier et al. \cite{Maier2021} extended this framework to 3D Cardiac CT imaging, utilizing a spatial transformer module to estimate motion vector fields from partial angles, enabling more effective motion compensation. In cardiac CT, however, motion is primarily driven by non-rigid movements, such as those caused by the heart's rhythmic activity. Notably, HR-pQCT scans are conducted on peripheral sites using a thin slice, necessitating a motion model different from those used for cardiac movements. \textcolor{black}{To the best of our knowledge, no existing literature has concurrently addressed both motion modeling and solving rigid motion artifacts, e.g., cortical streaking and trabecular smearing due to motion in HR-pQCT bone imaging.}

% This work introduces a novel sinogram-based patient motion model for peripheral sites that can simulate realistic HR-pQCT rigid motion artifacts from motion score 1 (artifact-free) data. Subsequently, these pairs are trained with an Edge-enhanced Self-attention Wasserstein Generative Adversarial Network (ESWGAN) with gradient penalty (ESWGAN-GP) to solve motion artifacts. The novel contribution of the network lies in its integration of an edge-enhancing generator, and a content loss function, which collectively enable effective preservation of the trabecular structures during motion correction in both the source domain (simulated data) and the target domain (real-world patient motion data). Additionally, incorporating an attention block enhances the model's ability to capture long-range dependencies within the cortical bone architectures, i.e., leverage information from the non-corrupted cortical bones to further contribute to its robustness in addressing motion artifacts. 
\textcolor{black}{The proposed method introduces a motion correction framework that integrates an \textcolor{black}{adapted} motion simulation algorithm to model in-plane motion. This algorithm generates motion artifacts from motion-free data, creating paired datasets. These datasets are subsequently employed to train an Edge-enhanced Self-attention Wasserstein Generative Adversarial Network with Gradient Penalty (ESWGAN-GP), designed to mitigate motion artifacts effectively.}
% The network's key innovation lies in its integration of an edge-enhancing generator and a content loss function, which together preserve trabecular structures during motion correction in both simulated and real-world patient motion data. Additionally, the inclusion of an attention block improves the model’s ability to capture long-range dependencies within cortical bone architecture by leveraging information from unaffected cortical regions, enhancing robustness in artifact correction. 
The key contributions of this work are outlined below:
\begin{enumerate}
\item A motion simulation model is \textcolor{black}{utilized}, which is based on the rotation of the 2D object to be imaged. This method builds upon the principles of \textcolor{black}{in-plane single-step motion simulation}, wherein prior work \cite{sadik2024physics} has demonstrated that applying random \textcolor{black}{alterations} to the sinogram, followed by reconstruction using the Simultaneous Iterative Reconstruction Technique (SIRT) \cite{Gregor2008}, results in motion-corrupted images that resemble those observed in real-world scenarios.

\item Followed by the generation of motion-corrupted and ground truth image pairs, a motion correction method is proposed based on utilizing a Wasserstein Generative Adversarial Network with Gradient Penalty (WGAN-GP) as the backbone network\cite{10.5555/3295222.3295327}.

\item Self-attention networks in both the generator and the discriminator are utilized to capture a wide range of spatial features (SWGAN-GP).

\item A Sobel-kernel-based Convolutional Neural Network (SCNN) is integrated in addition to the skip connections in the U-Net generator to enhance the preservation of edges in the bone micro-structures (ESWGAN-GP).

\item Utilization of \textcolor{black}{VGG-based} content loss in conjunction with the adversarial loss to ensure robust reconstruction of the bone micro-structures.

\item \textcolor{black}{Two variations of the ESWGAN-GP model are also introduced: ESWGAN-GPv1, which incorporates both pixel-wise loss and Total Variation (TV) loss into the training objective, and ESWGAN-GPv2, which extends ESWGAN-GPv1 by integrating a U-Net shaped discriminator to further enhance performance.}

This collective approach constitutes a \textcolor{black}{comprehensive} motion correction pipeline specifically designed for \textcolor{black}{mitigating} rigid motion artifacts in HR-pQCT bone imaging. 
\end{enumerate}

\section{Problem Formulation}\label{sec2}
In this section, we conceptualize the motion correction problem as both a motion correction and a de-blurring problem. 
% Although HR-pQCT involves 3D acquisition, it is performed on a thin slice, leading us to assume that the motion parameters remain consistent across all slices within a volume.
Let $\bold{f}$ $\epsilon$  $\mathbb{R}^{N_{w} \times N_{h}}$ denotes the object to be imaged which has a width and height of $N_{w}$ and $N_{h}$. The corresponding sinogram can be represented by $\bold{s}$ $\epsilon$  $\mathbb{R}^{N_{\rho} \times N_{\theta}}$, where $N_{\rho}$ is the number of projection lines and $N_{\theta}$ is the number of projection angles used while imaging the object. The HR-pQCT acquisition can be described by the following objective function shown in Equation \ref{eq:1}.

\begin{equation}\label{eq:1}
{
 \bold{s}=\mathcal{R}\bold{f} + \bold{\varepsilon}
}
\end{equation}

Here, $\mathcal{R}: \mathbb{R}^{N_{w} \times N_{h}} \rightarrow \mathbb{R}^{N_{\rho} \times N_{\theta}}$ represents the Radon transform matrix, when multiplied with the input image $\bold{f}$, it outputs the sinogram of the image, $\bold{s}$, and $\varepsilon$ denotes the system noise. However, patient motion will cause the ideal object $\bold{f}$ to be degraded to $\bold{f^{*}} = \mathcal{\bold{M}}\bold{f}$, retaining the following Equation \ref{eq:2}

\begin{equation}\label{eq:2}
{
 \bold{s^{'}}=\mathcal{R}\mathcal{\bold{M}}\bold{f} + \varepsilon
}
\end{equation}
where $\mathcal{\bold{M}}$ denotes the motion matrix, and $\bold{s^{'}}$ denotes the corrupted sinogram.

The motion-corrupted image can be acquired by applying the inverse Radon transform as indicated in Equation \ref{eq:3}.

\begin{equation}\label{eq:3}
{
 \bold{f^{*}}=\mathcal{R}^{-1}\bold{s^{'}}
}
\end{equation}

However, achieving a sharp and fully converged reconstruction requires numerous iterative steps. Reducing the number of iterations leads to a non-ideal solution, $\bold{f^{'}}$ shown in Equation \ref{eq:4}

\begin{equation}\label{eq:4}
{
 \bold{f^{'}}=\bold{f^{*}} + \delta
}
\end{equation}
where $\delta$ denotes a marginal blurring effect resulting from the loss of high-frequency information due to a reduction in the number of iterations. Thus, utilizing fewer iterations during the reconstruction of the corrupted sinogram makes this problem analogous to motion correction as well as a de-blurring problem similar to \cite{8536420}. The motion corrupted and blurred image, $\bold{f^{'}}$, and the corresponding ground truth, $\bold{f}$ will be utilized for training the GAN.

% \begin{figure}[t]
%     \centering
%     \includegraphics[width=0.9\linewidth]{FST.jpg}
%     \caption{Illustration of the Fourier slice theorem. (a) A sinogram of a motion free HR-pQCT image; (b) $k$-space of the corresponding image; (c) $k$-space of the motion corrupted (rotated by $\alpha$) image. \textcolor{black}{Identitical k-spaces are used to demonstrate the Fourier slice theorem.}}
%     \label{fig:FST}
% \end{figure}

\subsection{Simulation of Motion}\label{subsec3}
\begin{figure}[htp]
    \centering
    \includegraphics[width=0.9\linewidth]{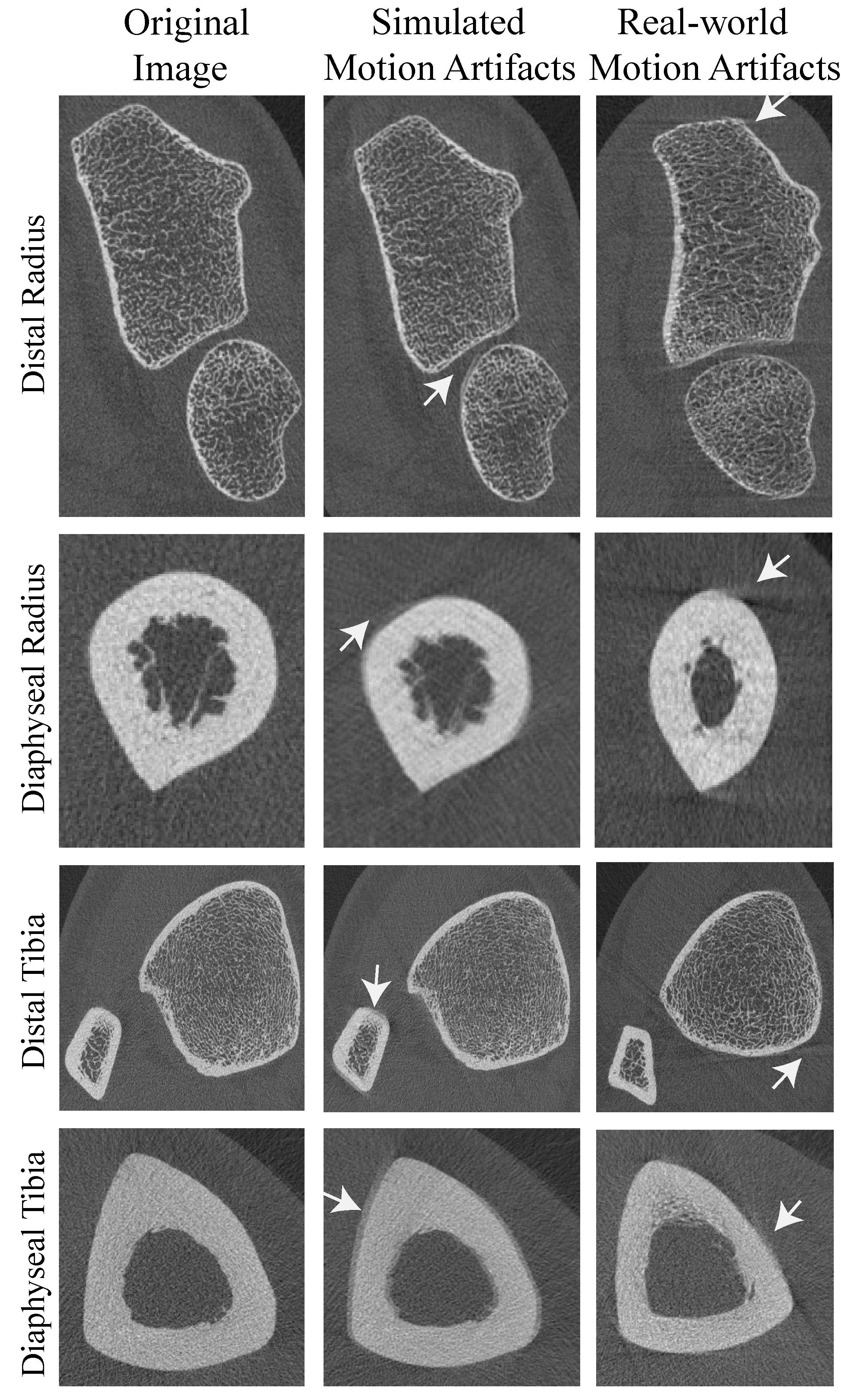}
    \caption{\textcolor{black}{Motion artifacts are simulated utilizing the sinogram-based in-plane motion simulation framework.} The first two columns display the motion score 1 (motion-free image) alongside the corresponding simulated motion artifacts in the same participant, while the third column represents a real-world motion scenario in a different participant, \textcolor{black}{which is reconstructed using the scanner's default con-beam reconstruction.} The arrows denote the particular artifacts that this study aims to address. The Region of Interest (ROI) is cropped to display only the bone area in the figure.}
    \label{fig:3}
\end{figure}

\textcolor{black}{A single-step rotation within the reconstruction plane was simulated by replacing a selected range of consecutive rows in the sinogram with corresponding rows from the sinogram of a rotated version of the object. This procedure results in a composite sinogram, where a portion of the projections corresponds to the object in one orientation and the remaining projections correspond to a different orientation, effectively mimicking a sudden change in pose during data acquisition. The corrupted sinogram was reconstructed using the SIRT algorithm with deliberately reduced iteration counts, resulting in a marginally blurred motion-corrupted image.} The outcome of the proposed simulation method is illustrated in Fig. \ref{fig:3}, where four peripheral sites are exposed to various degrees of motion simulation and are compared with real-world motion-corrupted data from the scanner.

\subsection{Proposed Model for Motion Correction}\label{subsec4}
Following the motion simulation scheme, we obtained 483 pairs of simulated motion-corrupted and ground-truth data from four different peripheral sites. Each pair consists of 168 ground-truth images and 168 simulated motion-corrupted images. The ROI was extracted from these image pairs, resized to 256 × 256, and subsequently processed by the proposed ESWGAN-GP model, which maps the motion-corrupted images to their corresponding ground truth, effectively correcting the motion artifacts in the process.

\textcolor{black}{It is important to note that the proposed ESWGAN-GP model operates on two-dimensional (2D) images, meaning it has been trained on individual 2D slices. Consequently, to address motion artifacts in a three-dimensional (3D) volume, each 2D slice must be processed separately. The rationale for adopting a slice-wise 2D approach is based on the characteristics of HR-pQCT imaging, which acquires exceptionally thin slices (approximately 61 $\mu$m). This high in-plane resolution reduces the variation in motion artifacts across the entire stack of 168 slices, justifying the use of a 2D correction method. However, due to cone beam geometry, motion artifacts may still vary across slices. Nevertheless, as the ESWGAN-GP is trained slice-by-slice, it will perform corrections in a slice-wise manner, rendering the potential variation in motion artifacts across slices less impactful.}

During the training, the predicted results are compared with the ground-truth to calculate the loss, which is subsequently back-propagated to update the network parameters. Specifically, the motion-corrupted image, $\bold{f}^{'}$ undergoes a deep learning network parameterized with $G_{\theta}$ to predict a motion-free image $\bold{\hat{f}}$ as shown in equation \ref{eq:12}.

\begin{equation}\label{eq:12}
\bold{\hat {f}} = G_{\theta}(\bold{f^{'}}).
\end{equation}

 An objective function or loss is minimized to obtain the optimal $\hat{\theta}$, ensuring that $G_{\hat{\theta}}$ predicts a motion-free image, $\bold{\hat{f}}$ that closely approximates the corresponding ground-truth, $\bold{f}$. 
\begin{equation}\label{eq:13}
\hat{\theta} = \arg\min_{\theta} \sum_{i} \mathcal{L}(\hat{f}_i(\theta), f_i)
\end{equation}

The backbone of the proposed model is the Generative Adversarial Network (GAN), which consists of a generator network $G$ and a discriminator network $D$. The generator network $G$ maps the input space $F'$ to the output space $F$, i.e.,
\[
G: F' \rightarrow F(G), \quad \text{where } \bold{f'} \in F', \ \bold{f} \in F.
\]
The generator's task is to predict a motion-compensated image $\bold{\hat{f}}$ from the input image $\bold{f'}$, while the discriminator $D$ attempts to distinguish between the generated motion-compensated image $\bold{\hat{f}}$ and the ground truth image $\bold{f}$. Training continues until the generator is able to fool the discriminator, making it unable to differentiate between the generated motion-compensated image and the ground truth motion-free image. The following optimization problem is addressed during the training process.

\begin{equation}\label{eq:13}
\min_{G} \max_{D} \mathcal{L}_{\text{GAN}}(G, D)
\end{equation}

%Furthermore, content loss is employed to ensure that the generator adheres strictly to %the characteristics of the training data i.e., maintaining consistency with the input. 
The subsequent sections are organized as follows, Section 1 provides a detailed explanation of the adversarial loss, while Section 2 offers a concise overview of the content loss.

\subsubsection{Adversarial Loss}

Unlike the negative log-likelihood used in traditional GANs \cite{Goodfellow2020}, we employ the earth-mover distance or Wasserstein GAN (WGAN) to ensure differentiability with respect to the input, thereby mitigating the training difficulties associated with log-likelihood loss. In particular, the discriminator's gradient with respect to the input is optimized more effectively in WGANs than in standard GANs, leading to improved performance of the generator. Further improvements can be achieved by integrating Gradient Penalty into WGAN (WGAN-GP), which modifies the optimization problem to the following form.

\begin{align}\label{eqn:14}
\min_G \max_{D} \mathcal{L}_{\text{WGAN-GP}}(G, D) &=
\mathbb{E}_{\mathbf{f'}}[D(G(\mathbf{f'}))] - \mathbb{E}_{\mathbf{f}}[D(\mathbf{f})] \nonumber \\
&\quad + \lambda \mathbb{E}_{\tilde{\mathbf{f}}}\left[(\|\nabla_{\tilde{\mathbf{f}}} D(\tilde{\mathbf{f}})\|_2 - 1)^2\right]
\end{align}

Here, \(E(.)\) represents the expectation operator. The first two terms correspond to the discriminator's criteria, where both the generated image and the corresponding ground truth image are input into the discriminator, and the Wasserstein distance is measured. Conversely, the third term represents the gradient penalty, which enforces 1-Lipschtiz constraint i.e., the gradients of the discriminator output with respect to input are normalized and constrained to remain below 1; any deviation beyond this threshold incurs a penalty.

 \begin{figure*}[htp]
    \centering
    \includegraphics[width=0.9\linewidth]{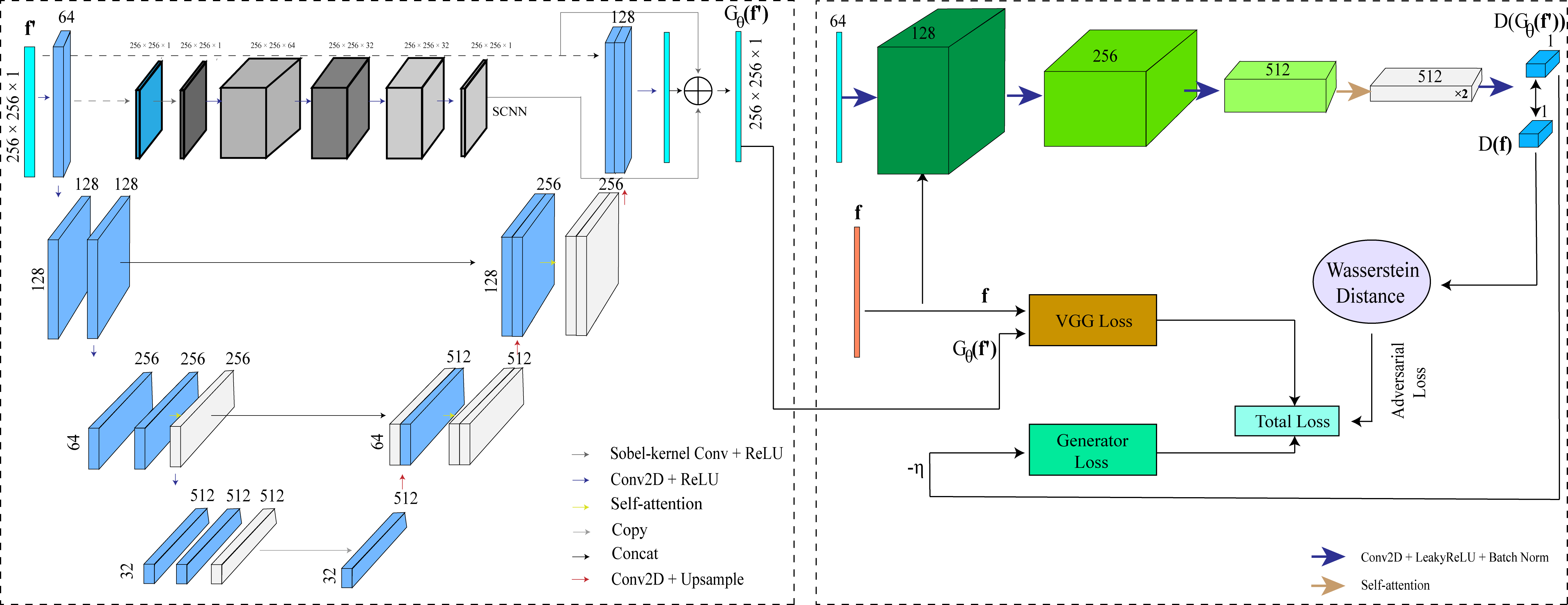}
    \caption{Illustration of the proposed ESWGAN-GP network, which consists of a generator and a discriminator. The motion-corrupted image, $\mathbf{f'}$, is fed into the \textbf{\textit{generator}}, which encodes it into a feature space with dimensions $32 \times 32 \times 512$. This encoded feature is then passed through the decoder, which reconstructs it back to the original input image dimensions (256). In the final step, the SCNN's input and output are combined element-wise with the decoded feature to generate the predicted motion-compensated image, $G_\theta(\mathbf{f'})$. In SCNN, the input image's edges are extracted using a Sobel-kernel convolution and processed through a series of convolutional and activation layers to produce feature maps with the same spatial dimensions as the input image. The predicted image, $G_\theta(\mathbf{f'})$ is passed through the \textbf{\textit{discriminator}} alongside its motion-free counterpart, $\mathbf{f}$, and a Wasserstein distance is utilized to minimize the distance. The dotted lines indicate connections originating from the preceding layer. This figure is most effectively viewed in a digital format. }
    \label{fig:6}
\end{figure*}

\subsubsection{Content Loss}

In addition to the generator loss, a VGG-based perceptual \(l_{1}\)-loss is employed \cite{8099502}, where both the ground truth image and the generated image are processed through a VGG network, which is pre-trained on a natural image dataset, Imagenet \cite{5206848}. 
%This approach is analogous to qualitative evaluation by the human visual system, with the VGG network serving as a proxy for human perceptual assessment.
The content loss function is shown in Equation \ref{eq:15}.
\begin{equation}\label{eq:15}
\begin{split}
 \mathcal{L}_{\text{content}}(G) = \eta \times -\mathbb{E}(D(G(\bold{f'}))) \\
 + \mathbb{E}_{(f', f)} \left[\| VGG(G(\bold{f'})) - VGG(\bold{f}) \|_1 \right]
\end{split}
\end{equation}

In addition to the adversarial loss, the final optimization objective for this problem is formulated as follows:

\begin{equation}\label{eqn:16}
\begin{split}
\min_G \max_{D} \mathcal{L}_{\text{WGAN-GP}}(D, G) = & \; \mathbb{E}_{\mathbf{f'}}[D(G(\mathbf{f'}))] - \mathbb{E}_{\mathbf{f}}[D(\mathbf{f})] \\
& + \lambda \mathbb{E}_{\tilde{\mathbf{f}}}\left[(\|\nabla_{\tilde{\mathbf{f}}} D(\tilde{\mathbf{f}})\|_2 - 1)^2\right] \\
& + \mathcal{L}_{\text{content}}
\end{split}
\end{equation}

\subsection{The Proposed Network Architecture}

The backbone of the generator is a U-Net architecture, consisting of an encoder block that employs multiple convolutional layers to transform the input into feature space, and a decoder block that reconstructs the images from the lower-level feature information \cite{10.1007/978-3-319-24574-4_28}. Skip connections are implemented between various layers of the encoder and decoder to prevent the vanishing gradient problem and maintain consistency with the original input. In addition to the skip connections originally proposed in the U-Net architecture, we introduce an edge enhancer block that transfers edge information directly from the input to the final output. This architecture effectively captures both low-level details through skip connections and global structure through downsampling and upsampling operations, while the additional edge enhancement block helps preserve cortical and trabecular edge features. Moreover, we utilized a self-attention module within both the generator and the discriminator to effectively capture long-range dependencies within an image. Incorporating attention blocks enables the evaluation of an image based on its global context, as opposed to the localized focus typically employed by conventional CNNs \cite{9124791}. The proposed network architecture is shown in Fig. \ref{fig:6}. The subsequent sections provide detailed descriptions of the edge enhancer block and the attention network.

% Conventional convolution kernels, such as \(3 \times 3\) or \(5 \times 5\), are designed to activate only local neighboring pixels, thereby failing to capture long-range spatial features or the global context of the entire image. Furthermore, incorporating attention blocks within the discriminator enables the evaluation of an image based on its global context, as opposed to the localized focus typically employed by conventional CNNs \cite{9124791}.

\subsubsection {Edge-enhancer Block}

A Sobel-kernel-based Convolutional Neural Network (SCNN) module is incorporated alongside the skip connection layer in the uppermost layer of the U-Net-based generator for edge enhancement. In SCNN, the input image undergoes Sobel kernel filtering \cite{5054605} to generate an edge-detected image. This edge-detected image is then subjected to a sequence of three consecutive convolutional layers followed by rectified Linear Units (ReLU) before being combined through matrix addition with the final output of the decoder (SCNN in Fig. \ref{fig:6}).
To derive the edge image from the input, the Sobel kernel employs two distinct filters oriented along the horizontal and vertical directions. Convolution with these filters is followed by calculating the magnitude squared image, which results in the final edge-detected image. The rationale for incorporating an edge enhancement module into the deep learning network is grounded in the high-resolution nature of HR-pQCT, which necessitates precise edge reconstruction to ensure measurement of cortical porosity and thickness as well as trabeculae thickness \cite{Cheung2013}.

% \begin{figure}[t]
%     \centering
%     \includegraphics[width=0.9\linewidth]{Figure 4.jpg}
%     \caption{Illustration of the proposed Edge-Enhancer block, referred to as the Sobel Convolutional Neural Network (SCNN). In this block, the input image's edges are extracted using a Sobel-kernel convolution. The output is then processed through a series of convolutional and activation layers to produce feature maps with the same spatial dimensions as the input image.}
%     \label{fig:4}
% \end{figure}

% \[
% G_x = \begin{bmatrix}
% -1 & 0 & 1 \\
% -2 & 0 & 2 \\
% -1 & 0 & 1
% \end{bmatrix}
% ;
% \quad
% G_y = \begin{bmatrix}
% -1 & -2 & -1 \\
% 0 & 0 & 0 \\
% 1 & 2 & 1
% \end{bmatrix}
% \]

% \begin{figure*}[htp]
%     \centering
%     \includegraphics[width=0.9\linewidth]{ATTMAP.jpg}
%     \caption{An illustration of a single self-attention block is presented. The feature maps from the previous layer, denoted as \( \bold{x} \), are passed through three \( 1 \times 1 \) convolution layers to generate the query (\( f(\bold{x}) \)), key (\( g(\bold{x}) \)), and value (\( h(\bold{x}) \)) feature maps. The transpose of the query is multiplied by the keys to produce an attention map, which is then multiplied by the value feature maps. The query and key determine the relevant portions of the values to which attention should be applied. A final \( 1 \times 1 \) convolution layer is applied to generate the output feature maps, which are then added to the input (not shown in the figure).}
%     \label{fig:5}
% \end{figure*}

\subsubsection {Self-attention Block}

The proposed approach incorporates self-attention blocks into specific layers of the generator and discriminator (white blocks shown in Fig. \ref{fig:6}) to effectively capture long-range dependencies within an image. The attention block, in the generator, is connected directly to the feature maps produced by lower convolutional layers rather than to a flattened output from a fully connected layer, or the upper layers in the generator. This design choice is intentional: by working with 2D feature maps, the attention block can selectively focus on certain regions or patterns within the spatial structure of the image. This setup enables the attention mechanism to emphasize important parts of the image while maintaining the original spatial relationships between pixels, which would be lost if the data were flattened or data were still in the low-level feature state, e.g., in the upper portion of the generator. As a result, the attention block can better highlight relevant features and patterns within the 2D layout of the image \cite{pmlr-v37-xuc15,pmlr-v97-zhang19d}. 

\section{Experimental Details and Results}

In this section, we describe the dataset employed in this study, followed by a detailed explanation of the experimental setup, including both the motion simulation parameters and the ESWGAN-GP parameters. Subsequently, an ablation study is conducted to assess the contribution of each component of the ESWGAN-GP network. Finally, the proposed motion correction scheme is validated using both a simulated motion dataset (source) and a real-world motion-corrupted dataset (target). For quantitative evaluation, \textcolor{black}{peak-signal-to-noise ratio (PSNR), structural
similarity index measure (SSIM) \cite{5596999}, and visual information
fidelity (VIF)} \cite{Sheikh2006, 1532311} metrics are used on the simulated dataset where ground truth data is available. 
These metrics are also used for assessing the reconstruction performance on the target domain. The target domain input, which has reduced iteration during reconstruction with inherent motion artifacts, is compared with their original raw image through these metrics to measure data fidelity—specifically, the retention of micro-structural information despite motion correction.
\subsection{Dataset}
Between 2018 and 2022, a cohort of 558 volunteers who underwent HR-pQCT scanning \textcolor{black}{in} the \textcolor{black}{Function, Imaging, and Testing (FIT) Resource Core} of the Indiana Center for Musculoskeletal Health’s Clinical Research Center (Indianapolis, Indiana)\footnote{https://medicine.iu.edu/research-centers/musculoskeletal/clinical-research/for-participants/fit-core} met the criteria for initial inclusion in this study, with selection being random and including motion grades of all types, 1 to 5. The FIT Core received Institutional Review Board (IRB) approval from Indiana University, and each participant provided written informed consent prior to imaging.
% Initial recruitment to the FIT Core included both research participants from ongoing musculoskeletal studies seeking standardized outcomes and members of the local community. Inclusion criteria encompassed volunteers aged 18 years or older. Participants were required to be physically able to comply with study protocols and remain seated during scanning. Exclusion criteria included pregnancy, breastfeeding, active implanted medical devices (e.g., pacemakers, defibrillators), and malignancies requiring chemotherapy. No exclusions were applied based on race, ethnicity, or gender.  
HR-pQCT scans (XtremeCT II, Scanco Medical, Bruttisellen, Switzerland) \textcolor{black}{were acquired on the non-dominant arm at 4\% (Distal Radius) and 30\% (Diaphyseal Radius) proximal from a distal radius reference line, and the contralateral leg at 7.3\% (Distal Tibia) and 30\% (Diaphyseal Tibia) proximal from a distal tibia reference line}, as we have previously described \cite{Warden2021}. Participants were positioned supine on a movable treatment plinth, with the limb of interest stabilized using padded carbon fiber casts provided by the manufacturer. Participants were instructed to remain motionless during scanning. Scanning parameters included 68 kVp and 1.47 mA, with 168 slices (covering 10.2 mm of bone) acquired at a voxel size of 60.7 $\mu$m. Scanner stability was maintained by routinely scanning phantoms with density and volume inserts as per the manufacturer's guidelines.
Motion was assessed by a trained operator using a visual grading score (VGS) \cite{Sode2011}, ranging from 1 (no motion artifacts) to 5 (severe streaking, cortical disruptions, and trabecular blurring). Table \ref{tab:dataset_summary} provides a comprehensive summary of participants and the corresponding images used in the study.

\begin{table}[h!]
    \centering
    \renewcommand{\arraystretch}{1.4}
    \begin{tabular}{|>{\centering\arraybackslash}p{2.9 cm}|>{\centering\arraybackslash}p{1.8cm}|>{\centering\arraybackslash}p{1.4cm}|>{\centering\arraybackslash}p{1.2cm}|}
        \hline
        \textbf{Bone Type} & \textbf{Train-test Split} & \textbf{Participants} & \textbf{Images} \\
        \hline
        \multirow{3}{*}{Distal (4$\%$) Radius} & Train (Source) & 90 & 15,120 \\
        \cline{2-4}
                                       & Test (Source) & 13 & 2,184 \\
        \cline{2-4}
                                       & Test (Target) & 40 & 6,720 \\
        \hline
        \multirow{3}{*}{\textcolor{black}{Diaphyseal} (30$\%$) Radius} & Train (Source) & 100 & 16,800 \\
        \cline{2-4}
                                         & Test (Source) & 30 & 5,040 \\
        \cline{2-4}
                                         & Test (Target) & 13 & 2,184 \\
        \hline
        \multirow{3}{*}{Distal (7.3$\%$) Tibia} & Train (Source) & 90 & 15,120 \\
        \cline{2-4}
                                      & Test (Source) & 36 & 6,048 \\
        \cline{2-4}
                                      & Test (Target) & 14 & 2,352 \\
        \hline
        \multirow{3}{*}{\textcolor{black}{Diaphyseal} (30$\%$) Tibia} & Train (Source) & 90 & 15,120 \\
        \cline{2-4}
                                        & Test (Source) & 34 & 5,712 \\
        \cline{2-4}
                                        & Test (Target) & 8 & 1,344 \\
        \hline
        \multicolumn{2}{|r|}{\textbf{Total}} & 558 & 93,744 \\
        \hline
    \end{tabular}
    \caption{Summary of the dataset utilized in this study. The \textbf{\textit{source}} data consists of VGS 1 images used for simulating motion artifacts, thereby containing ground truth information. Conversely, the \textbf{\textit{target}} data comprises real-world motion-corrupted images of VGS - 2, 3, 4, and 5. Images from the same participants are strictly segregated between the training and testing sets, avoiding any patient overlap in the train-test split across all experiments.}
    \label{tab:dataset_summary}
\end{table}

\subsection{Experimental Details}

\subsubsection{Motion Simulation Parameters}

The sinogram-based motion simulation proposed in Section~\ref{subsec3} involves two key variables: \textcolor{black}{the rotation angle of the imaging object, and the number of lines altered in the sinogram due to motion}. The rotation angles are selected from a range of values, specifically from \(-\frac{\pi}{20}\) to \(\frac{\pi}{120}\). \textcolor{black}{The rationale for this assumption is grounded in prior findings indicating that substantial rotational movements are uncommon in peripheral anatomical sites \cite{Pauchard2011}}. Out of the 1800 total projection lines, 200 consecutive projections were altered to introduce motion artifacts.
\textcolor{black}{To process the target dataset, which comprises real-world motion-corrupted data, the images are input into the simulation framework without the addition of synthetic motion. This approach aims to generate comparable marginal blurring effects to those observed in the source dataset.}  
% These selections are not the only possible options; however, a comprehensive parameter study is beyond the scope of this research. It is hypothesized that a broader variation in parameters could generate a wider range of motion-corrupted images, potentially leading to more robust motion correction. 
The motion simulation was implemented using MATLAB 2023a\footnote{https://www.mathworks.com/products/new\_products/release2023a.html} with the help of ASTRA Toolbox \cite{vanAarle:16, VANAARLE201535, PMID:21840398}\footnote{https://astra-toolbox.com/}.
\subsubsection{Neural Network Parameters}

In the proposed ESWGAN-GP, the penalty term in the discriminator is governed by the parameter \(\lambda\), which manages the balance between the Wasserstein loss and the gradient penalty. For all experiments, \(\lambda\) is set to 0.2. Additionally, the ADAM \cite{KingBa15} optimizer is employed, with decay rates set to \(\beta_{1} = 0.5\) and \(\beta_{2} = 0.999\). The learning rate is fixed at \(8 \times 10^{-5}\). The batch size for all experiments is set to 1. The trade-off between the generator loss and VGG-based perceptual loss is controlled by the parameter,
\( \eta\) = \(1 \times 10^{-3}\). All the codes for the neural network were implemented using the PyTorch framework\footnote{https://pytorch.org/}, and the simulations were conducted on an NVIDIA RTX A5500 GPU with 24 GB of memory. \textcolor{black}{The codes are available at:} \href{https://github.com/fsa125/HR-pQCT-Motion-Correction---ESWGAN-GP}{\textcolor{blue}{https://github.com/fsa125/HR-pQCT-Motion-Correction---ESWGAN-GP}}.

\begin{table*}[htp]
\centering
\caption{Comparison of PSNR, SSIM, and VIF values across different sites for \textcolor{black}{seven} different models.}
\resizebox{\linewidth}{!}{
\begin{tabular}{lcccccccccccc}
\hline
& \multicolumn{3}{c}{Distal Radius} & \multicolumn{3}{c}{\textcolor{black}{Diaphyseal} Radius} & \multicolumn{3}{c}{Distal Tibia} & \multicolumn{3}{c}{\textcolor{black}{Diaphyseal} Tibia} \\
\cline{2-13}
& PSNR & SSIM & VIF & PSNR & SSIM & VIF & PSNR & SSIM & VIF & PSNR & SSIM & VIF \\
\hline
WGAN (S) & $25.15 \pm 1.24$ & $0.78 \pm 0.02$ & $0.58 \pm 0.04$ & $25.36 \pm 1.16$ & $0.71 \pm 0.03$ & $0.59 \pm 0.05$ & $24.53 \pm 0.98$ & $0.79 \pm 0.02$ & $0.59 \pm 0.03$ & $25.66 \pm 1.10$ & $0.74 \pm 0.03$ & $0.53 \pm 0.04$ \\
WGAN (T) & $28.62 \pm 1.07$ & $0.86 \pm 0.01$ & $0.64 \pm 0.03$ & $29.00 \pm 0.76$ & $0.81 \pm 0.01$ & $0.59 \pm 0.04$ & $27.15 \pm 0.84$ & $0.86 \pm 0.02$ & $0.65 \pm 0.03$ & $27.71 \pm 0.53$ & $0.81 \pm 0.01$ & $0.59 \pm 0.03$ \\
WGAN-GP (S) & $\textcolor{black}{25.99} \pm \textcolor{black}{1.27}$ & $\textcolor{black}{0.81} \pm \textcolor{black}{0.02}$ & $\textcolor{black}{0.70} \pm \textcolor{black}{0.04}$ & $25.82 \pm 0.97$ & $0.71 \pm 0.02$ & $0.75 \pm 0.04$ & $25.15 \pm 0.85$ & $0.81 \pm 0.02$ & $0.68 \pm 0.03$ & $26.58 \pm 0.87$ & $0.77 \pm 0.01$ & $0.69 \pm 0.03$ \\
WGAN-GP (T) & $\textcolor{black}{29.12} \pm \textcolor{black}{1.12}$ & $\textcolor{black}{0.87} \pm \textcolor{black}{0.01}$ & $\textcolor{black}{0.76} \pm \textcolor{black}{0.03}$ & $28.63 \pm 0.74$ & $0.79 \pm 0.03$ & $0.78 \pm 0.03$ & $27.36 \pm 0.75$ & $0.87 \pm 0.01$ & $0.73 \pm 0.03$ & $27.92 \pm 0.59$ & $0.82 \pm 0.01$ & $0.74 \pm 0.02$ \\
SWGAN-GP (S) & $\textcolor{black}{25.79} \pm \textcolor{black}{1.20}$ & $\textcolor{black}{0.78} \pm \textcolor{black}{0.02}$ & $\textcolor{black}{0.67} \pm \textcolor{black}{0.04}$ & $27.02 \pm 0.98$ & $0.76 \pm 0.02$ & $0.82 \pm 0.03$ & $24.85 \pm 0.97$ & $0.81 \pm 0.02$ & $0.70 \pm 0.03$ & $\textcolor{black}{26.35} \pm \textcolor{black}{1.18}$ & $\textcolor{black}{0.77} \pm \textcolor{black}{0.02}$ & $\textcolor{black}{0.71} \pm \textcolor{black}{0.03}$ \\
SWGAN-GP (T) & $\textcolor{black}{28.73} \pm \textcolor{black}{1.03}$ & $\textcolor{black}{0.85} \pm \textcolor{black}{0.01}$ & $\textcolor{black}{0.74} \pm \textcolor{black}{0.03}$ & $29.73 \pm 0.63$ & $0.82 \pm 0.01$ & $0.82 \pm 0.02$ & $27.00 \pm 0.88$ & $0.86 \pm 0.01$ & $0.74 \pm 0.03$ & $\textcolor{black}{28.22} \pm \textcolor{black}{0.55}$ & $\textcolor{black}{0.83} \pm \textcolor{black}{0.01}$ & $\textcolor{black}{0.77} \pm \textcolor{black}{0.02}$ \\
ESWGAN-GP (S) & $26.37 \pm 1.13$ & $0.81 \pm 0.02$ & $0.71 \pm 0.04$ & $27.13 \pm 1.00$ & $0.78 \pm 0.02$ & $0.82 \pm 0.03$ & $26.45 \pm 0.88$ & $0.86 \pm 0.02$ & $0.75 \pm 0.03$ & $27.18 \pm 0.88$ & $0.80 \pm 0.02$ & $0.75 \pm 0.03$ \\
ESWGAN-GP (T) & $ 29.23 \pm 1.00$ & $0.87 \pm 0.01$ & $0.77 \pm 0.03$ & $29.98 \pm 0.58$ & $0.84 \pm 0.00$ & $\bold{0.84} \pm \bold{0.02}$ & $29.17 \pm 0.76$ & $0.91 \pm 0.01$ & $0.80 \pm 0.03$ & $28.84 \pm 0.57$ & $0.85 \pm 0.00$ & $\bold{0.81} \pm \bold{0.02}$ \\
%\rowcolor{red!10}
ESWGAN-GPv1 (S) & $27.64 \pm 1.10$ & $0.85 \pm 0.01$ & $0.73 \pm 0.04$ & $24.65 \pm 0.95$ & $0.59 \pm 0.02$ & $0.60 \pm 0.05$ & $26.73 \pm 0.87$ & $0.86 \pm 0.01$ & $0.75 \pm 0.03$ & $24.68 \pm 0.93$ & $0.60 \pm 0.02$ & $0.56 \pm 0.04$ \\
%\rowcolor{red!10}
ESWGAN-GPv1 (T) & $ \bold{31.10} \pm \bold{0.93}$ & $\bold{0.90} \pm \bold{0.02}$ & $0.78 \pm 0.02$ & $26.99 \pm 0.77$ & $0.64 \pm 0.03$ & $0.60 \pm 0.03$ & $\bold{29.64} \pm \bold{0.77}$ & $\bold{0.91} \pm \bold{0.01}$ & $\bold{0.81} \pm \bold{0.02}$ & $26.00 \pm 0.59$ & $0.66 \pm 0.01$ & $0.61 \pm 0.03$ \\

%\rowcolor{red!10}
ESWGAN-GPv2 (S) & $27.44 \pm 1.10$ & $0.85 \pm 0.01$ & $0.73 \pm 0.03$ & $27.88 \pm 0.95$ & $0.81 \pm 0.02$ & $0.79 \pm 0.03$ & $26.60 \pm 0.89$ & $0.86 \pm 0.01$ & $0.74 \pm 0.03$ & $27.56 \pm 0.82$ & $0.80 \pm 0.02$ & $0.72 \pm 0.03$ \\
%\rowcolor{red!10}
ESWGAN-GPv2 (T) & $ 30.86 \pm 0.96$ & $0.90 \pm 0.01$ & $\bold{0.79} \pm \bold{0.03}$ & $\bold{31.12} \pm \bold{0.62}$ & $\bold{0.87} \pm \bold{0.01}$ & $0.80 \pm 0.02$ & $29.41 \pm 0.79$ & $0.91 \pm 0.01$ & $0.78 \pm 0.02$ & $\bold{29.29} \pm \bold{0.54}$ & $\bold{0.86} \pm \bold{0.01}$ & $0.77 \pm 0.02$ \\

GAN-CIRCLE (S) & $24.18 \pm 1.32$ & $0.73 \pm 0.04$ & $0.55 \pm 0.04$ & $24.27 \pm 1.37$ & $0.73 \pm 0.03$ & $0.59 \pm 0.05$ & $23.85 \pm 0.97$ & $0.73 \pm 0.03$ & $0.58 \pm 0.03$ & $24.90 \pm 1.36$ & $0.71 \pm 0.03$ & $0.53 \pm 0.04$ \\
GAN-CIRCLE (T) & $27.45 \pm 1.09$ & $0.81 \pm 0.02$ & $0.61 \pm 0.04$ & $28.49 \pm 0.72$ & $0.82 \pm 0.02$ & $0.59 \pm 0.04$ & $26.34 \pm 0.84$ & $0.80 \pm 0.02$ & $0.63 \pm 0.03$ & $27.40 \pm 0.53$ & $0.78 \pm 0.01$ & $0.59 \pm 0.03$ \\
\hline
\end{tabular}
}
\label{table:1}
\end{table*}

\begin{table*}[h]
\centering
\caption{Statistical analysis using ANOVA and Tukey-Kramer for PSNR, SSIM, and VIF scores across different sites in the source dataset. A $\checkmark$ indicates statistical significance at p-value $\leq$ 0.01.}
\resizebox{\linewidth}{!}{
\begin{tabular}{lcccccccccccc}
\hline
& \multicolumn{3}{c}{Distal Radius} & \multicolumn{3}{c}{\textcolor{black}{Diaphyseal} Radius} & \multicolumn{3}{c}{Distal Tibia} & \multicolumn{3}{c}{\textcolor{black}{Diaphyseal} Tibia} \\
\cline{2-13}
& PSNR & SSIM & VIF & PSNR & SSIM & VIF & PSNR & SSIM & VIF & PSNR & SSIM & VIF \\
\hline
WGAN vs WGAN-GP & \textcolor{black}{\checkmark} & \textcolor{black}{\checkmark}  & \textcolor{black}{\checkmark}  & \checkmark   & \checkmark  & \checkmark   & \checkmark  & \checkmark  & \checkmark  & \checkmark  & \checkmark  & \checkmark \\
WGAN vs SWGAN-GP & \textcolor{black}{\checkmark}  & \textcolor{black}{\checkmark}  & \textcolor{black}{\checkmark}  & \checkmark   & \checkmark  & \checkmark   & \checkmark  & \checkmark  & \checkmark  & \textcolor{black}{\checkmark}  & \textcolor{black}{\checkmark}  & \textcolor{black}{\checkmark} \\
WGAN vs ESWGAN-GP* & \checkmark  & \checkmark  & \checkmark  & \checkmark   & \checkmark  & \checkmark   & \checkmark  & \checkmark  & \checkmark  & \checkmark  & \checkmark  & \checkmark \\
WGAN-GP vs SWGAN-GP & \textcolor{black}{\checkmark}  & \textcolor{black}{\checkmark} & \textcolor{black}{\checkmark}  & \checkmark   & \checkmark  & \checkmark   & \checkmark  & \ding{55}  & \checkmark  & \checkmark & \checkmark  & \checkmark \\
WGAN-GP vs ESWGAN-GP* & \checkmark  & \textcolor{black}{\ding{55}}  & \checkmark  & \checkmark   & \checkmark  & \checkmark   & \checkmark   & \checkmark  & \checkmark  & \checkmark  & \checkmark  & \checkmark \\
SWGAN-GP vs ESWGAN-GP* & \checkmark   & \checkmark  &  \checkmark & \checkmark   & \checkmark  & \checkmark  & \checkmark  & \checkmark  & \checkmark  & \checkmark  & \checkmark  & \checkmark \\

GAN-CIRCLE vs ESWGAN-GP* & \checkmark   & \checkmark  &  \checkmark & \checkmark   & \checkmark  & \checkmark  & \checkmark  & \checkmark  & \checkmark  & \checkmark  & \checkmark  & \checkmark \\

\hline
\end{tabular}
}
\label{table:3}
\end{table*}

\begin{figure*}[htb]

    \includegraphics[width=0.99\linewidth]{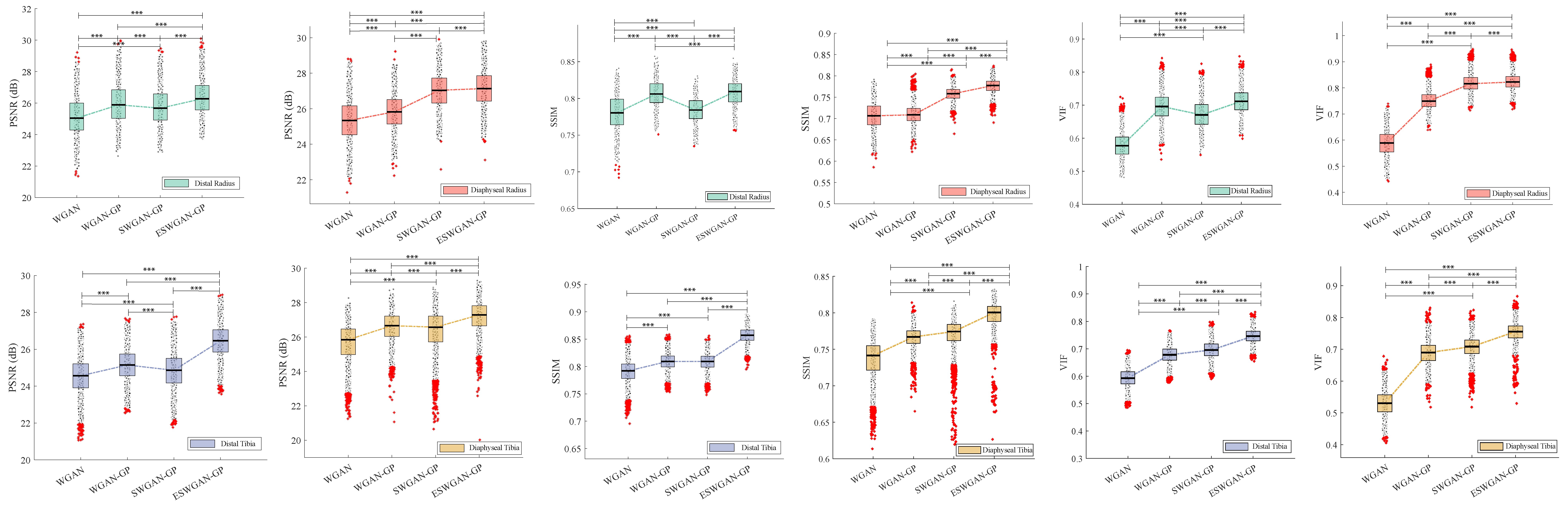}
    \caption{Comparing the PSNR, SSIM, and VIF values of four models (WGAN, WGAN-GP, SWGAN-GP, and ESWGAN-GP) across four anatomical sites (Distal Radius, \textcolor{black}{Diaphyseal} Radius, Distal Tibia, and \textcolor{black}{Diaphyseal} Tibia) in the source dataset. Each plot represents the performance of the four models at a specific site. Red stars indicate outliers in the data. \textcolor{black}{The outliers are calculated based on interquartile range (IQR). IQR is the distance between the third quartile (Q3), and the first quartile (Q1). Any datapoint less than Q1 - 1.5$\times$IQR is considered a lower outlier, and greater than Q3 + 1.5$\times$IQR is considered an upper outlier.} The symbol `***' denotes a p-value of $\leq 0.01$, as determined by a paired t-test under the assumption of unequal variance. Pairs that are not connected indicate a lack of statistically significant difference.} 
    \label{fig:13}
\end{figure*}

\subsection{Results}
%The results section is organized into two %distinct parts. 
%Initially, 

% Next, we compare the proposed method with two well-known GANs for super-resolution, namely GAN-CIRCLE \cite{Guha2020, You2020} and SACNN \cite{Li2020}.
% Secondly, we present findings derived from the extensive dataset and conduct both qualitative and quantitative evaluations using the aforementioned evaluation metrics.
% \begin{figure}[t]

%     \includegraphics[width=0.99\linewidth]{Figure7.jpg}
%     \caption{Experimental results on the source (simulated) dataset. Arrows highlight the wisp artifacts that require correction. The rightmost column displays the absolute difference between the ground truth image and the motion-compensated image, with the color bar providing a quantitative measure of pixel-level absolute differences.}
%     \label{fig:7}
% \end{figure}

% \begin{figure}[t]

%     \includegraphics[width=0.99\linewidth]{Figure9.jpg}
%     \caption{Experimental results on the target (real-world motion corrupted) dataset. Arrows highlight the wisp artifacts that require correction. The rightmost column displays the absolute difference between the ground truth image and the motion-compensated image, with the color bar providing a quantitative measure of pixel-level absolute differences.In principle, there is no ground truth; however, the input target images are processed through the motion simulation step without inducing any motion, in order to prevent distribution shift, as explained in Section \ref{discussion}.}
%     \label{fig:9}
% \end{figure}

 We first evaluate the selection of the proposed network, ESWGAN-GP by comparing it to its preceding models: 1) WGAN, 2) WGAN-GP, and 3) SWGAN-GP. Consistency in the content loss has been maintained across all networks, incorporating both the generator loss and the VGG-based perceptual loss. For the ablation study, all four peripheral sites are utilized, and tests are conducted on simulated, and target data. PSNR, SSIM, and VIF values, along with their corresponding standard deviations, are calculated. \textcolor{black}{Means and standard deviations of the performance metrics are computed over the entire test set to assess model performance over the whole dataset.} \textcolor{black}{Additionally, two variants of the ESWGAN-GP model were evaluated: ESWGAN-GPv1, which incorporates pixel-wise loss with weights 2 and total variation (TV) loss with weights 0.01, and ESWGAN-GPv2, which extends ESWGAN-GPv1 by integrating a U-Net-shaped discriminator architecture, similar to Real-ESRGAN \cite{9607421}.} Secondly, we compared the performance of the proposed ESWGAN-GP with GAN-CIRCLE \cite{Guha2021,You2020}, a previously applied method for CT super-resolution, after making slight modifications to adapt it to our proposed dataset for performance evaluation. Due to the unavailability of code for the only existing motion correction abstract \cite{steiner2022correction} we opted to implement GAN-CIRCLE as the most suitable alternative for comparison.
 \subsubsection{Ablation Study} \label{ablation}
 From Table \ref{table:1}, the superior performance of ESWGAN-GP and its variants is seen across the four anatomical sites in the source (S), and target (T) dataset. \textcolor{black}{Specifically, within the source domain, ESWGAN-GPv1 (S) yields the highest PSNR values at distal anatomical sites, while ESWGAN-GPv2 (S) demonstrates superior PSNR performance at the \textcolor{black}{diaphyseal} site. A similar trend can also be observed in the target domain.} \textcolor{black}{An observable trend is that the quantitative metric values in the target domain are higher than those in the source domain. This can be attributed to the fact that the target images are processed through the motion simulation framework without the addition of synthetic motion, thereby preserving their inherent motion artifacts characterized by relatively mild blurring. In contrast, the source domain includes images with artificially introduced motion, particularly more pronounced rotational motion, resulting in more severe artifacts. Consequently, the superior metric performance in the target domain is consistent with the comparatively lower degree of motion corruption.} Statistical analysis by ANOVA and Tukey-Kramer post hoc analysis indicate that ESWGAN-GP results in higher image quality than the other three models, shown in Table \ref{table:3}.
\textcolor{black}{Owing to its superior performance across all four peripheral sites (Fig.~\ref{fig:13}), ESWGAN-GP was selected for further analyses.}
 \begin{figure*}[htp]

    \includegraphics[width=0.99\linewidth]{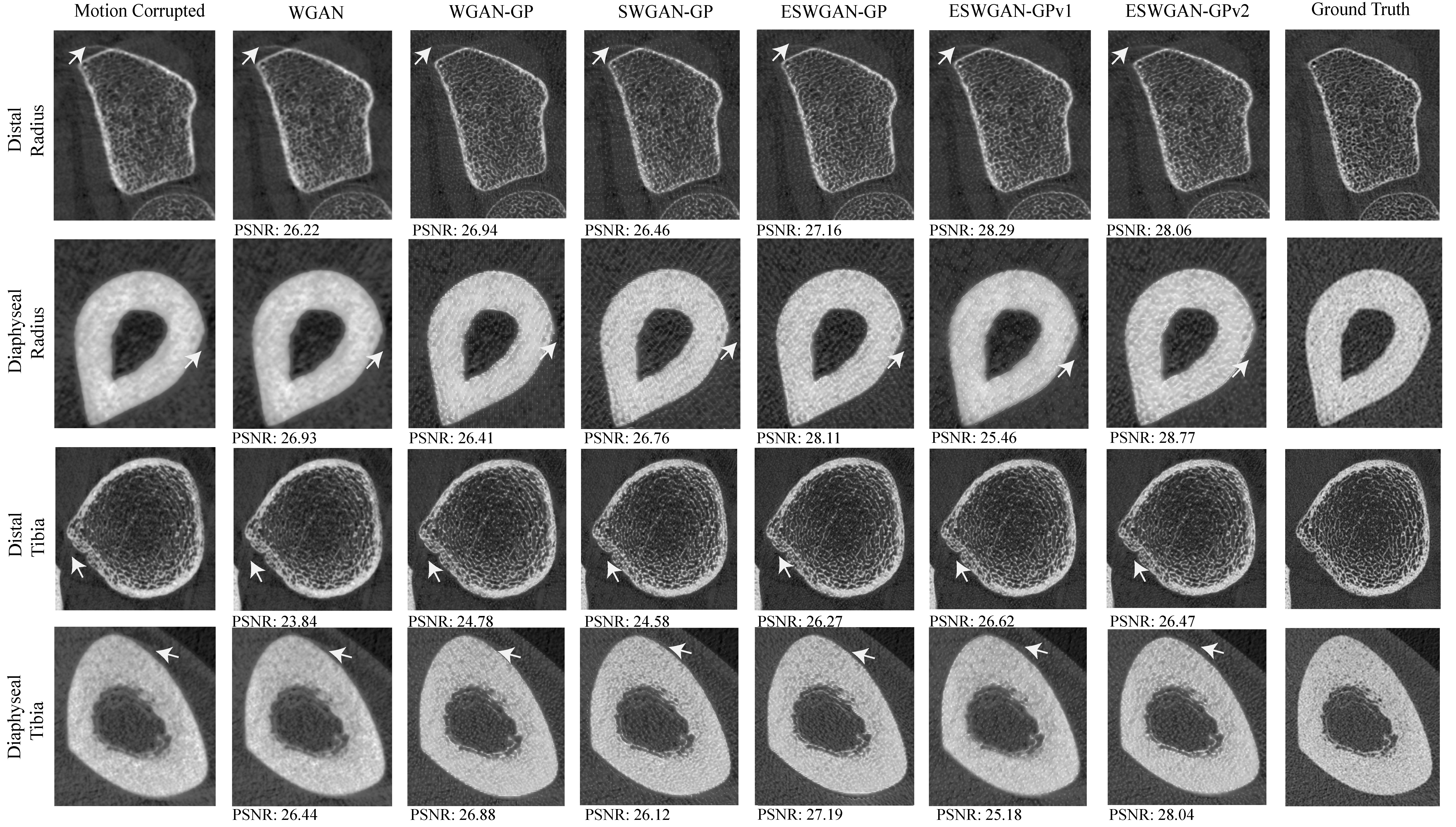}
    \caption{A qualitative assessment of the ablation study is presented. The leftmost column displays the input image \textcolor{black}{from the unseen source test dataset}, followed by the outputs generated by the proposed deep neural networks, and finally, the rightmost column presents the ground truth. The motion artifacts, highlighted by arrows, visually illustrate how these artifacts evolve by incorporating different network blocks. The corresponding difference image is not included in this figure.}
    \label{fig:14}
\end{figure*}
Fig. \ref{fig:14} presents a qualitative analysis from the ablation study, highlighting the progression of motion correction achieved through the \textcolor{black}{six} evaluated models. The figure reveals a consistent trend: the incorporation of a gradient penalty enhances network stability, leading to improved outcomes compared to WGANs at all four sites. Notably, the inclusion of the self-attention block in SWGAN-GP results in better cortical bone reconstruction, particularly in the \textcolor{black}{Diaphyseal} Radius and Tibia. Furthermore, the final enhancement comes from the edge enhancer block (SCNN), which significantly improves the reconstruction of fine microstructural details, most notably in the Distal Radius and Tibia. One observed limitation of ESWGAN-GP is the trabecularization of cortical bone, particularly pronounced in \textcolor{black}{diaphyseal} regions where cortical thickness is greater. To mitigate this issue, we implemented pixel-wise loss and TV loss in ESWGAN-GPv1, and incorporated a U-Net-shaped discriminator in ESWGAN-GPv2. As illustrated in Fig. \ref{fig:14}, these modifications improved trabecularization artifacts in distal regions; however, residual artifacts remain visible in the \textcolor{black}{diaphyseal} sites. \textcolor{black}{Although the variants of ESWGAN-GP demonstrated superior performance in reducing motion artifacts both qualitatively (Fig. \ref{fig:14}) and quantitatively (Table \ref{table:1}) at specific anatomical sites, we selected the original ESWGAN-GP for further analysis of bone geometry due to its consistent performance across all sites.}

\begin{figure*}[htp]

    \includegraphics[width=0.99\linewidth]{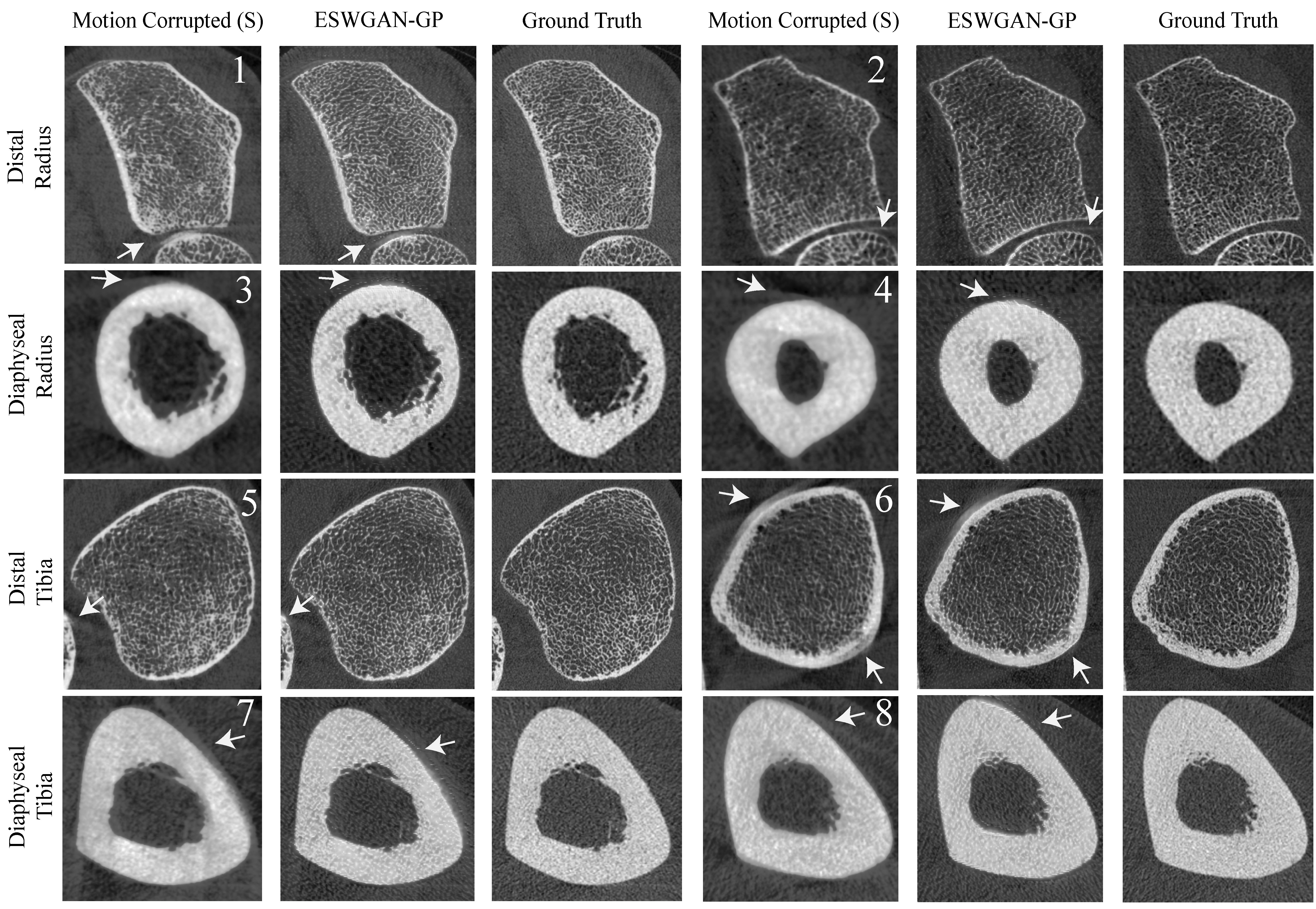}
    \caption{\textcolor{black}{Representative heavy motion-corrupted samples from the unseen source test dataset processed by ESWGAN-GP. Arrows highlight regions exhibiting motion artifacts. The corresponding difference images are not shown in this figure.}}
    \label{fig:review2}
\end{figure*}

\textcolor{black}{Fig. \ref{fig:review2} illustrates the qualitative performance of ESWGAN-GP on source domain images affected by severe motion artifacts. In the first example from the Distal Radius (``1" in Fig. \ref{fig:review2}), two distinct cortical disruptions are evident—one located in the superior region and the other in the inferior region. ESWGAN-GP substantially restores the superior cortical break, while in the inferior region, the cortical boundary becomes discernible, albeit with some residual wisps remaining. The second example from the Distal Tibia (``6" in Fig. \ref{fig:review2}) depicts an atypical case of motion artifacts characterized by a rotation angle larger than typically encountered in practical scenarios. ESWGAN-GP successfully reconstructs the cortical boundary at the inferior surface; however, minor streak artifacts persist in the superior region, though the boundary is notably more distinct compared to the motion-corrupted input. A similar pattern is observed across all other examples, further demonstrating the robustness of ESWGAN-GP in correcting varying degrees of motion artifacts.}

 \subsubsection{Comparison with GAN-CIRCLE}

\begin{figure*}[htp]

    \includegraphics[width=0.99\linewidth]{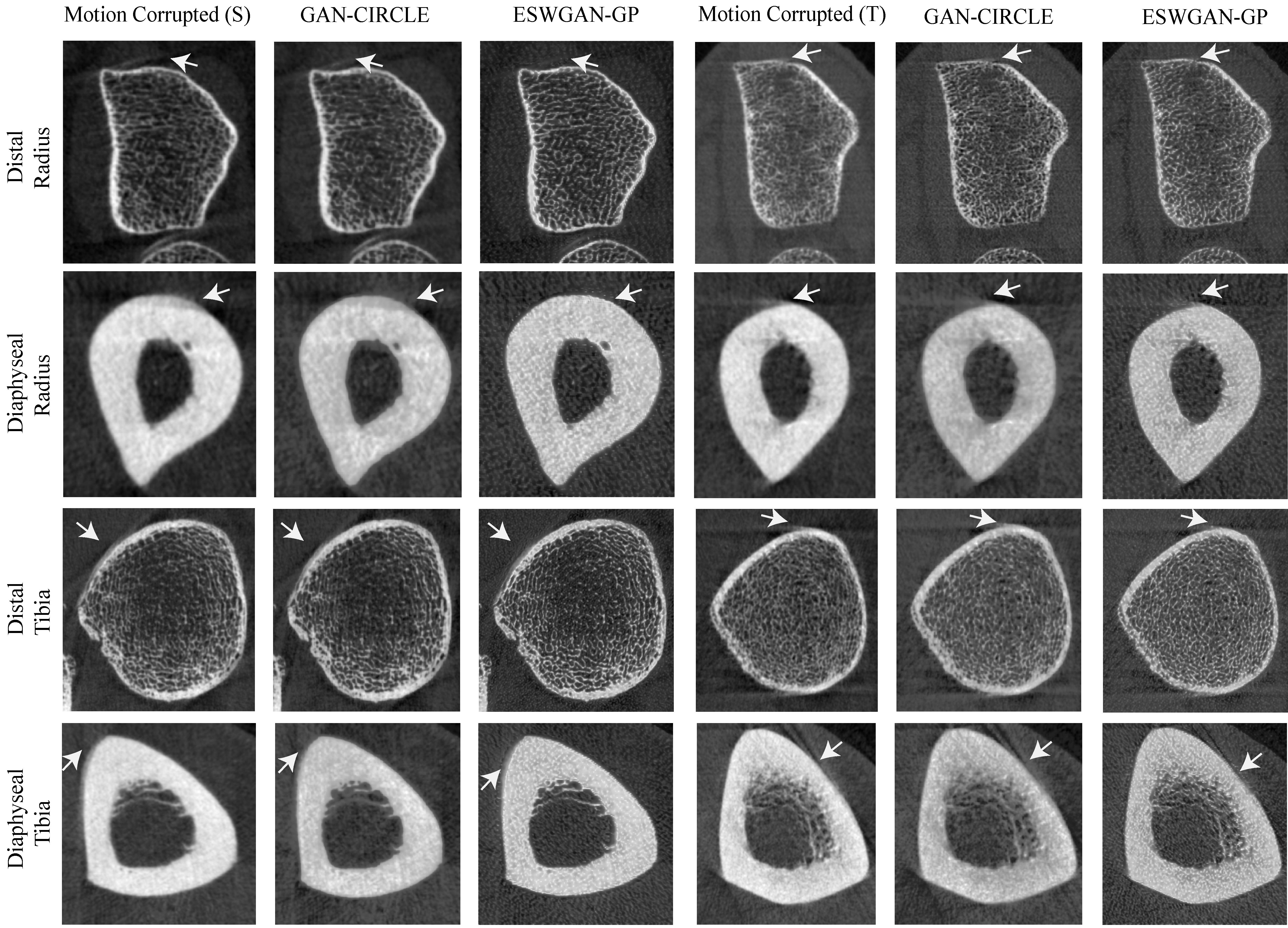}
    \caption{The proposed ESWGAN-GP is compared with GAN-CIRCLE in both the simulated \textcolor{black}{unseen test} data and the target \textcolor{black}{unseen test} data. The motion artifacts, highlighted by arrows, visually illustrate how these artifacts are resolved in the two networks. The corresponding difference images are not included in this figure.}
    \label{fig:15}
\end{figure*}

As seen in Table \ref{table:1}, ESWGAN-GP outperforms GAN-CIRCLE by approximately 2.5 dB in PSNR on the source dataset and demonstrates an average improvement of 1.8 dB on the target dataset. Statistical analysis in Table \ref{table:3} further claims that the mean metric values of the ESWGAN-GP outperform GAN-CIRCLE. 
% A qualitative analysis of both the source and target data in Fig. \ref{fig:15} shows that ESWGAN-GP more accurately resolves cortical streaks compared to GAN-CIRCLE. 
This can be attributed to the design choices in the two models. GAN-CIRCLE relies on cycle-consistency, which forces the generated data to closely resemble the source data, thereby preserving consistency with the input. However, resolving motion artifacts does not necessitate maintaining such strict input consistency. Fig. \ref{fig:15} illustrates the aforementioned phenomena: while GAN-CIRCLE enforces consistency, ESWGAN-GP leverages a self-attention mechanism that captures long-range dependencies, resulting in superior reconstruction of the cortical boundary in the generated images.

\section{Discussion}\label{discussion}

Owing to its capacity to deliver detailed evaluations of cortical and trabecular bone microstructure and mineralization, HR-pQCT is increasingly favored for fracture risk prediction in osteoporosis. It serves as a valuable complement to aBMD measurements obtained from DXA, while also enabling the detection of bone mineral alterations associated with conditions such as chronic kidney disease (CKD), thus providing a more comprehensive assessment of bone fragility \cite{Gazzotti2023}. \textcolor{black}{Moreover, HR-pQCT is increasingly being utilized in longitudinal studies, including those assessing the effects of bisphosphonates such as alendronate, risedronate, and ibandronate in managing bone-related disorders among postmenopausal women \cite{Burghardt2010, Bala2013, Chapurlat2012}.}
However, its performance can be severely hindered by motion artifacts, which remain a significant challenge. Repeated scans increase the time burden on both staff and patients and frequently fail to mitigate artifacts leading to compromised data quality. 

Although advancements in motion grading have been made, the lack of robust motion correction methods continues to obstruct accurate interpretation of key bone parameters. Motion-compromised images fail to represent true bone structure, thereby limiting the potential of HR-pQCT in clinical and research settings. Addressing this limitation is essential to enhance the reliability of HR-pQCT data and fully realize its promise as a non-invasive tool for bone health assessment.

In this study, we \textcolor{black}{optimize} \textcolor{black}{an in-plane} motion simulation model and \textcolor{black}{propose a deep learning-based} correction model for HR-pQCT bone imaging.
% The motion simulation was independently evaluated by a radiologist, and the qualitative findings indicate that the simulated artifacts closely resemble real-world motion artifacts observed in HR-pQCT imaging. 
\textcolor{black}{Specifically,} we employed the motion simulation model to generate pairs of motion-corrupted and motion-free (ground truth) images, which were then used for training within a supervised learning framework. \textcolor{black}{Experimental results indicate that the proposed motion correction framework can effectively reduce, though not entirely eliminate, motion artifacts in cortical bone and enhance the visualization of trabecular bone architecture. This study represents an initial step toward post-acquisition correction of motion artifacts in HR-pQCT.}

While the proposed study presents compelling results, it is not without limitations. Firstly, the simulation assumes parallel beam geometry when generating sinograms from images on a slice-by-slice basis, whereas HR-pQCT employs cone-beam geometry for sinogram acquisition. \textcolor{black}{This is a common simplification adopted by several studies \cite{5193053,Pauchard2011}. This presents a notable limitation, as HR-pQCT inherently utilizes cone-beam geometry for image acquisition. Nevertheless, this simplification affects only the simulation phase. The ESWGAN-GP model operates in the image domain and is designed to leverage spatial information between motion-corrupted and ground truth images, which provides a motion correction framework that is agnostic to the simulation model.} In the absence of literature detailing the HR-pQCT acquisition process, the authors opted to proceed with parallel beam geometry sinogram simulation, validating this approach by comparing the simulated motion artifacts to those in the original motion-corrupted images. Additionally, the number of angles required for image acquisition was too large to be accommodated within the current hardware capabilities. Future research will aim to increase the number of angles during sinogram generation to more closely align with the scanner's requirements. \textcolor{black}{Furthermore, in designing our simulation protocol, we followed the precedent set by Pauchard et al. \cite{Pauchard2011}, who applied rotational motion that ranges between -1.5$^{\circ}$ and 5$^{\circ}$. In our study, we extended this range to -9$^{\circ}$ (-$\pi$/20) to 1.5$^{\circ}$ ($\pi$/20) to explore a broader yet still plausible motion spectrum. Nevertheless, the assumption regarding this range may not fully capture all relevant scenarios. A thorough parameter study would indeed be necessary to rigorously determine optimal values and assess the robustness of the approach under varied motion conditions.}
%Additionally, the simulation was constrained to in-plane motion, excluding translation along the z-%axis. This assumption is based on the spatial limitations inherent to HR-pQCT systems, where %peripheral sites have restricted movement range, unlike full-body scanners used in CT or MRI where six degrees of freedom (DOF) motion is more prevalent. Did not finish}
% Furthermore, although raw sinograms from the scanner are available, there is limited information provided by SCANCO or found online regarding the algorithms used for their reconstruction. Consequently, to fully integrate the proposed method into the scanner system, these issues will need to be addressed in subsequent studies.

Two primary factors limit the realism of the simulation method. First, the image resolution has been significantly reduced from $2304 \times 2304$ pixels to $256 \times 256$ pixels due to computational limitations. Secondly, the proposed motion simulation algorithm prioritizes in-plane rotational motion, in contrast to the model developed by Pauchard et al. \cite{5193053,Pauchard2011}, which incorporates both in-plane (i.e., x-y plane) and longitudinal (i.e., z-axis) translational components. \textcolor{black}{This restricts the practical applicability of the proposed method. Nonetheless, in-plane translation can be incorporated by quantifying sinogram mismatches between parallelized projections, as described in \cite{Sode2011}, while z-axis translation can be simulated by substituting projection lines from adjacent slices, effectively modeling displacement along the acquisition axis. Since these translational components can be integrated within the existing framework, future efforts will aim to generate more realistic and diverse datasets, thereby improving the utility of data-driven motion correction strategies. Importantly, the proposed ESWGAN-GP architecture is agnostic to the specific type of motion simulation, enabling training on datasets derived from alternative modeling techniques. Future work will leverage this flexibility by combining simulation methods that include both the current approach and that of Pauchard et al. \cite{Pauchard2011}, and Sode et al. \cite{Sode2011} to develop a more generalizable and robust deep learning-based solution for motion correction.
}

\begin{figure}[t]

    \includegraphics[width=0.99\linewidth]{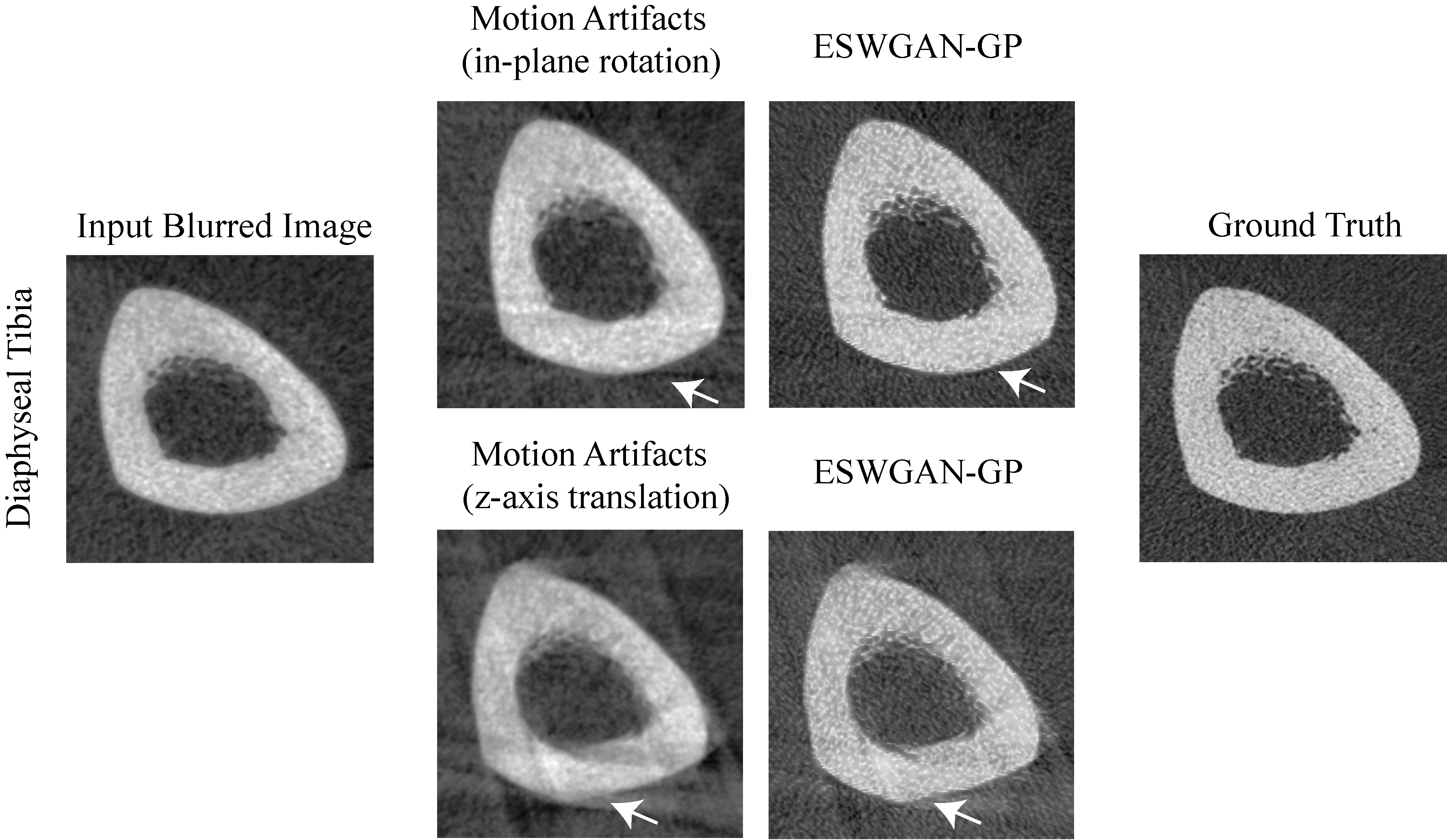}
    \caption{\textcolor{black}{Qualitative evaluation of the robustness of ESWGAN-GP. A volunteer was instructed to simulate in-plane rotational motion (top row) and z-axis translational motion (bottom row) during image acquisition. For ground truth reference, the same volunteer was scanned while being held securely to minimize motion. Corresponding difference images are not displayed in this figure.}}
    \label{fig:3review}
\end{figure}

\textcolor{black}{Nonetheless, the effectiveness of ESWGAN-GP in correcting motion artifacts resulting from translational motion along the z-axis was evaluated. As illustrated in Fig. \ref{fig:3review}, the model’s performance was assessed using artifacts generated through manual in-vivo simulation of both in-plane rotation and z-axis translation. The results demonstrate that ESWGAN-GP effectively restores cortical boundaries disrupted by in-plane rotation and exhibits a limited ability to correct motion artifacts resulting from z-axis translation. However, under conditions of substantial translational motion, as observed in the second row of Fig. \ref{fig:3review}, slight bending of the cortical boundary remains evident when compared to the ground truth. These findings underscore the need for a more robust and comprehensive simulation framework to enable effective correction of motion artifacts in HR-pQCT imaging.}

 \begin{figure}[htp]

    \includegraphics[width=0.99\linewidth]{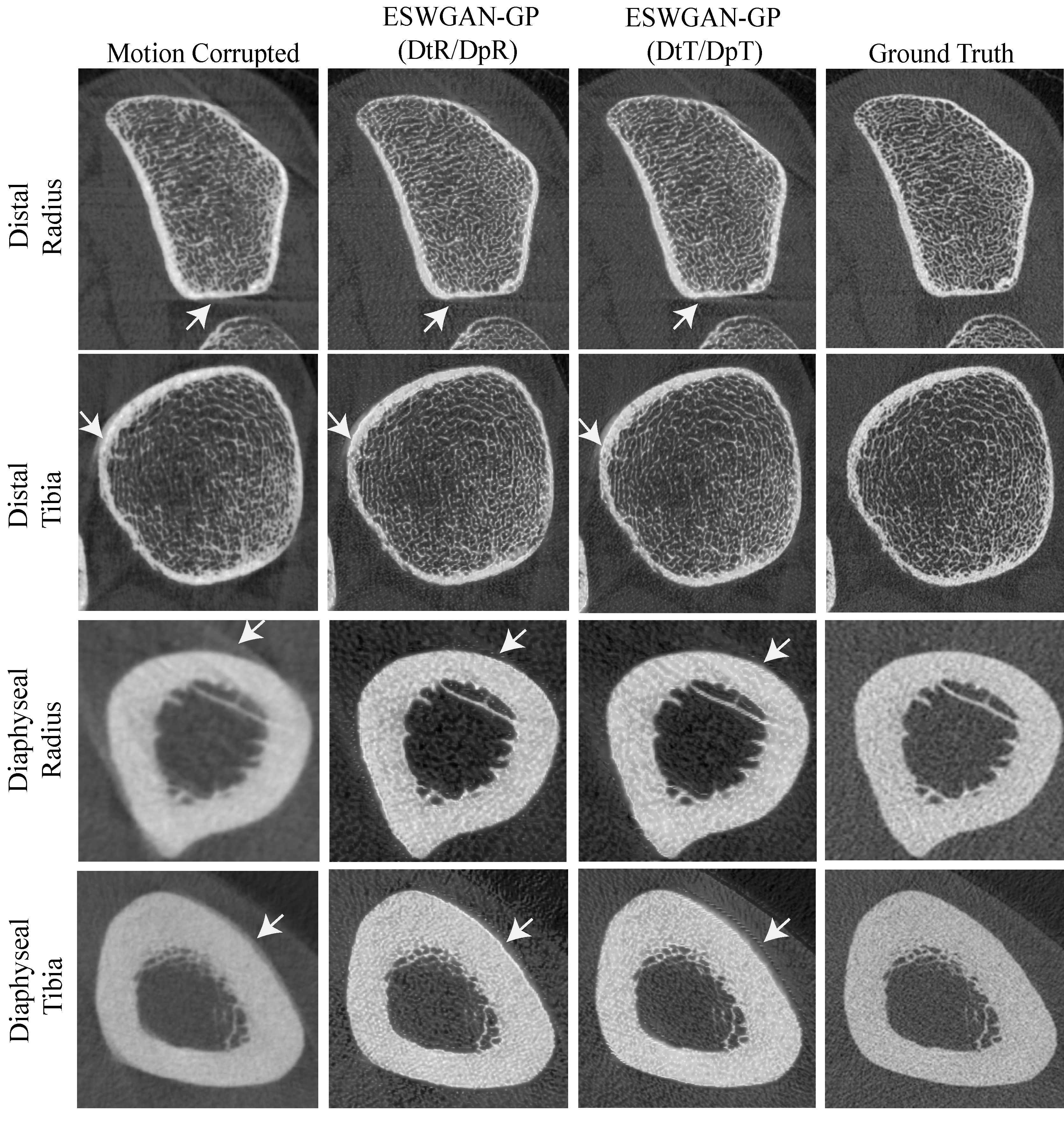}
    \caption{\textcolor{black}{A qualitative evaluation of the model generalizability in the unseen test data from the source domain is performed. ESWGAN-GP (DtR/DpR) indicates training on Distal or \textcolor{black}{Diaphyseal} Radius data, and ESWGAN-GP (DtT/DpT) on Distal or \textcolor{black}{Diaphyseal} Tibia data, with the first two rows representing distal sites and the last two \textcolor{black}{diaphyseal} sites. The white arrow highlights the artifacts targeted in this study; difference images are not shown.}}
    \label{fig:7review}
\end{figure}
In this study, the ESWGAN-GP model, along with its variants used in the ablation study, was trained on data from four distinct anatomical sites and evaluated on the corresponding sites. \textcolor{black}{Given the morphological similarities between the Distal Radius and Tibia, such as thin cortical bone and dense trabecular architecture, as well as between the \textcolor{black}{Diaphyseal} Radius and Tibia, where the cortical bone tends to be thicker with reduced trabecular density, we assessed model generalizability through cross-site evaluations. Specifically, the ESWGAN-GP model trained on the Radius was tested on the Tibia, and vice versa. The outcomes of this cross-site evaluation are presented in Table \ref{table:4}. 
}

\begin{table}[htp]
\centering
\caption{\textcolor{black}{Evaluation of model generalizability using PSNR, SSIM, and VIF metrics across cross-site training and testing on radius and tibia. The rows above the midline represent distal site experiments; those below represent \textcolor{black}{diaphyseal} site experiments.}}
\resizebox{\linewidth}{!}{
\begin{tabular}{lcccccc}
\hline
& \multicolumn{3}{c}{Distal/\textcolor{black}{Diaphyseal} Radius} & \multicolumn{3}{c}{Distal/\textcolor{black}{Diaphyseal} Tibia} \\
\cline{2-7}
& PSNR & SSIM & VIF & PSNR & SSIM & VIF \\
\hline

\textcolor{black}{ESWGAN-GP (DtR)} (S) & $26.37 \pm 1.13$ & $0.81 \pm 0.02$ & $0.71 \pm 0.04$ & $24.98 \pm 0.93$ & $0.81 \pm 0.02$ & $0.72 \pm 0.03$\\
\textcolor{black}{ESWGAN-GP (DtT)} (S) & $28.60 \pm 1.28$ & $0.83 \pm 0.02$ & $0.69 \pm 0.04$ & $26.45 \pm 0.88$ & $0.86 \pm 0.02$ & $0.75 \pm 0.03$ \\
\textcolor{black}{ESWGAN-GP (DtR)} (T) & $\bold{29.23} \pm \bold{1.00}$ & $\bold{0.87} \pm \bold{0.01}$ & $\bold{0.77} \pm \bold{0.03}$ & $27.30 \pm 0.81$ & $0.86 \pm 0.01$ & $0.78 \pm 0.03$\\
\textcolor{black}{ESWGAN-GP (DtT)} (T) & $24.98 \pm 0.93$ & $0.81 \pm 0.02$ & $0.72 \pm 0.03$ & $\bold{29.17} \pm \bold{0.76}$ & $\bold{0.91} \pm \bold{0.01}$ & $\bold{0.80} \pm \bold{0.03}$ \\
\hline
\textcolor{black}{ESWGAN-GP (DpR)} (S) & $27.13 \pm 1.00$ & $0.78 \pm 0.02$ & $0.82 \pm 0.03$ & $26.06 \pm 0.98$ & $0.75 \pm 0.02$ & $0.76 \pm 0.04$ \\

\textcolor{black}{ESWGAN-GP (DpT)} (S) & $26.61 \pm 1.01$ & $0.76 \pm 0.02$ & $0.74 \pm 0.03$ & $27.18 \pm 0.88$ & $0.80 \pm 0.02$ & $0.75 \pm 0.03$ \\

\textcolor{black}{ESWGAN-GP (DpR)} (T) & $\bold{29.98} \pm \bold{0.58}$ & $\bold{0.84} \pm \bold{0.00}$ & $\bold{0.84} \pm \bold{0.02}$ & $27.30 \pm 0.70$ & $0.80 \pm 0.02$ & $\bold{0.82} \pm {0.03}$ \\

\textcolor{black}{ESWGAN-GP (DpT)} (T) & $29.65 \pm 0.45$ & $0.83 \pm 0.02$ & $0.78 \pm 0.02$ & $\bold{28.84} \pm \bold{0.57}$ & $\bold{0.85} \pm \bold{0.00}$ & $0.81 \pm 0.02$ \\

\hline
\end{tabular}
}
\label{table:4}
\end{table}
As expected, models trained and evaluated on the same anatomical site—such as the ESWGAN-GP trained on Distal Radius (DtR) and tested on the same site demonstrated superior performance, with higher PSNR values, compared to scenarios where the model was tested on a different site, such as the Distal Tibia. Interestingly, when the ESWGAN-GP model trained on Distal Tibia (\textcolor{black}{ESWGAN-GP (DtT)}) was tested on Distal Radius, it yielded relatively high PSNR and SSIM values (28.60 and 0.83, respectively), although the VIF score was lower (0.69). The output in the target domain still falls within the expected behavior, wherein cross-site testing on the Distal Radius results in lower performance metrics compared to same-site testing (24.98 vs. 29.23 in PSNR). \textcolor{black}{The first two rows of} Fig. \ref{fig:7review} also exhibit the expected behavior where cross-testing performance is lower. Specifically, we can see in Distal Radius that the cortical breaks are better resolved in \textcolor{black}{\textcolor{black}{ESWGAN-GP (DtR)}} than in \textcolor{black}{ESWGAN-GP (DtT)}. However, \textcolor{black}{ESWGAN-GP (DtR)}, and \textcolor{black}{ESWGAN-GP (DtT)} both can marginally resolve cortical streaks in Distal Tibia. In both cases, ESWGAN-GP effectively does the deblurring task.

\textcolor{black}{A similar trend is observed in the \textcolor{black}{diaphyseal} regions, where same-site testing generally results in improved quantitative performance. An exception to this is the \textcolor{black}{ESWGAN-GP (DpT)} model, trained on the \textcolor{black}{Diaphyseal} Tibia and evaluated on the Radius, which achieves a higher PSNR than in the same-site testing (29.65 vs. 28.84) in the unseen target domain. Notably, the differences in fidelity metrics across sites are less pronounced in the \textcolor{black}{diaphyseal} regions compared to the distal regions. This can be attributed to the greater morphological similarity among \textcolor{black}{diaphyseal} sites. This observation is further supported by the last two rows of Fig. \ref{fig:7review}, where cortical streaking artifacts are more effectively mitigated in the \textcolor{black}{diaphyseal} regions, irrespective of the training site.} While our model demonstrates modest robustness in cross-site generalization, further investigations are warranted to assess the potential of deep learning approaches that exclude specific anatomical sites during training yet retain the ability to generalize effectively across sites.
\begin{table*}[t]
\centering
\caption{Segmentation performance metrics across four anatomical sites for different models evaluated on unseen source data.}
\resizebox{\linewidth}{!}{
\begin{tabular}{lcccccccccccc}
\hline
& \multicolumn{3}{c}{Distal Radius} & \multicolumn{3}{c}{\textcolor{black}{Diaphyseal} Radius} & \multicolumn{3}{c}{Distal Tibia} & \multicolumn{3}{c}{\textcolor{black}{Diaphyseal} Tibia} \\
\cline{2-13}
& Dice coefficient & Jaccard index & Hausdorff dist.   &Dice coefficient & Jaccard index & Hausdorff dist.  & Dice coefficient & Jaccard index & Hausdorff dist.  & Dice coefficient & Jaccard index & Hausdorff dist.  \\
\hline

Motion corrupted & $0.83 \pm 0.10$ & $0.72 \pm 0.13$ & $28.53 \pm 30.02$ & $0.98 \pm 0.01$ & $0.96 \pm 0.02$ & $6.95 \pm 4.12$ & $0.92 \pm 0.07$ & $0.85 \pm 0.09$ & $17.36 \pm 22.58$ & $0.98 \pm 0.01$ & $0.96 \pm 0.02$ & $9.70 \pm 4.50$ \\

WGAN & $0.83 \pm 0.10$ & $0.72 \pm 0.13$ & $28.57 \pm 29.96$ & $0.98 \pm 0.01$ & $0.96 \pm 0.02$ & $6.94 \pm 4.12$ & $0.92 \pm 0.07$ & $0.85 \pm 0.09$ & $17.32 \pm 22.50$ & $0.98 \pm 0.01$ & $0.96 \pm 0.02$ & $9.70 \pm 4.49$ \\

WGAN-GP & $0.87 \pm 0.08$ & $0.78 \pm 0.11$ & $23.90 \pm 25.23$ & $\bold{0.99} \pm \bold{0.01}$ & $0.97 \pm 0.01$ & $5.68 \pm 3.63$ & $0.93 \pm 0.05$ & $0.87 \pm 0.07$ & $\bold{14.26} \pm \bold{15.74}$ & $\bold{0.99} \pm \bold{0.01}$ & $\bold{0.97} \pm \bold{0.01}$ & $\bold{8.64} \pm \bold{4.50}$ \\

SWGAN-GP & $0.88 \pm 0.08$ & $\bold{0.79} \pm \bold{0.11}$ & $\bold{23.55} \pm \bold{27.00}$ & $0.99 \pm 0.00$ & $\bold{0.98} \pm \bold{0.01}$ & $5.54 \pm 4.00$ & $0.91 \pm 0.06$ & $0.84 \pm 0.08$ & $16.44 \pm 16.49$ & $0.98 \pm 0.01$ & $0.96 \pm 0.02$ & $9.70 \pm 4.49$ \\

ESWGAN-GP & $\bold{0.87} \pm \bold{0.09}$ & $0.78 \pm 0.12$ & $24.57 \pm 24.76$ & $0.99 \pm 0.00$ & $\bold{0.98} \pm \bold{0.01}$ & $\bold{5.08} \pm \bold{3.49}$ & $\bold{0.93} \pm \bold{0.06}$ & $\bold{0.87} \pm \bold{0.08}$ & $15.58 \pm 16.30$ & $0.98 \pm 0.01$ & $0.96 \pm 0.02$ & $9.45 \pm 4.71$ \\

\hline
\end{tabular}
}
\label{table:5}
\end{table*}

\begin{figure*}[t]

    \includegraphics[width=0.99\linewidth]{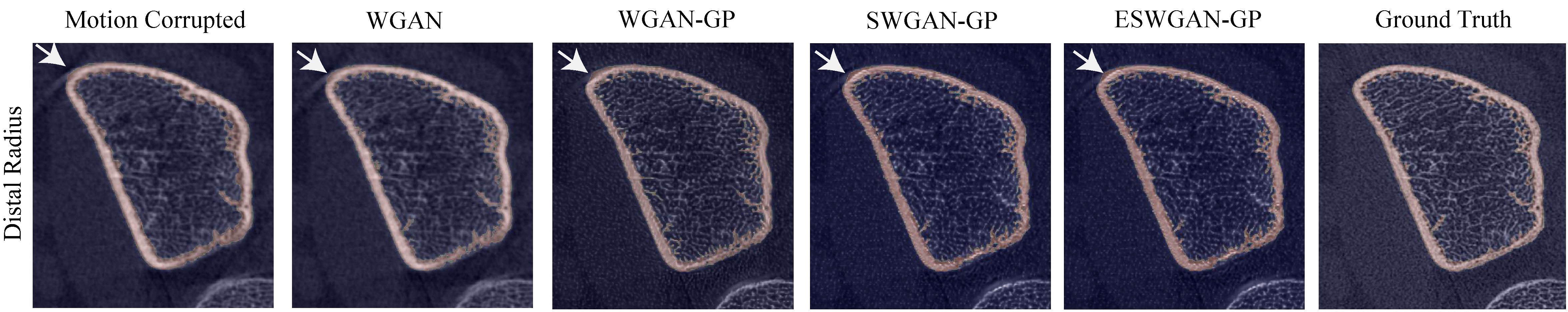}
    \caption{\textcolor{black}{Qualitative assessment of segmentation performance on motion-corrected images from the unseen source data generated by WGAN, WGAN-GP, SWGAN-GP, and ESWGAN-GP. The red boundary denotes the cortical bone segmented by autocontour, and the arrow highlights localized boundary refinement. For optimal clarity, view this figure on a digital display.}}
    \label{fig:10review}
\end{figure*}

\begin{figure*}[!htbp]

    \includegraphics[width=0.99\linewidth]{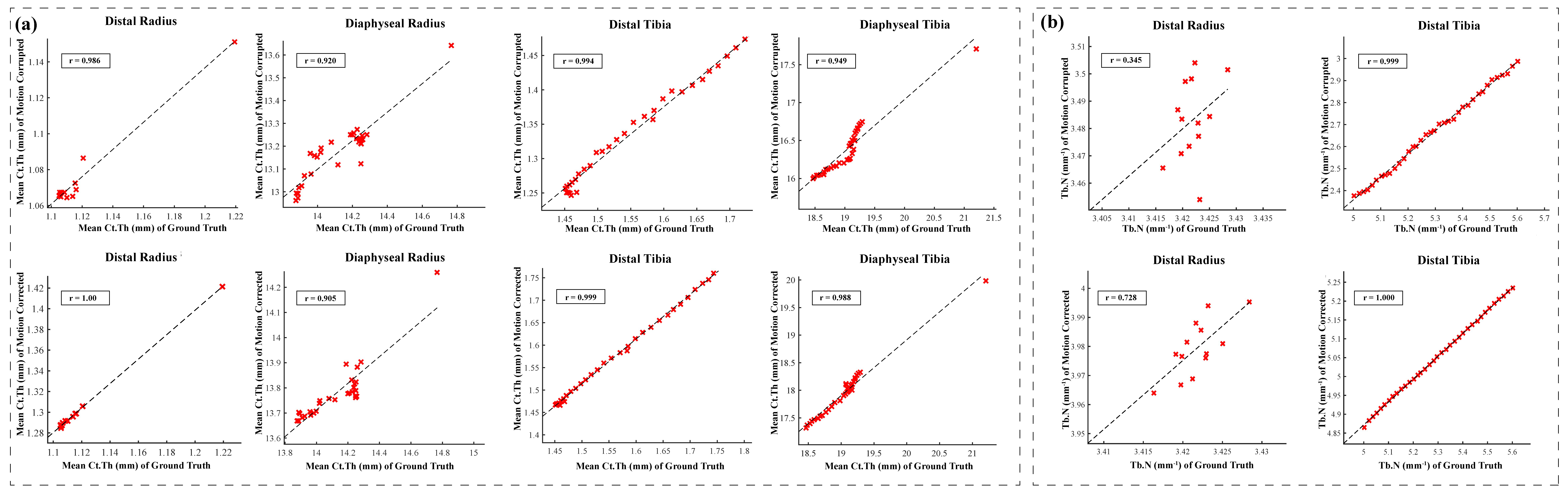}
    \caption{\textcolor{black}{Performance of ESWGAN-GP in restoring bone geometry in the unseen source dataset. (a) and (b) illustrate the effects of motion artifacts and their correction by ESWGAN-GP on cortical thickness (Ct.Th) and trabecular number (Tb.N), respectively. The first row shows measurements from motion-corrupted data, while the second presents those after correction. The Pearson correlation coefficient (r) quantifies how well motion correction preserves the underlying trends in the measurements.}}
    \label{fig:15review}
\end{figure*}

\textcolor{black}{HR-pQCT is primarily valued for its ability to provide quantitative assessments, including cortical thickness, trabecular number, and bone mineral density \cite{Gazzotti2023}. Accurate derivation of these metrics necessitates the segmentation of HR-pQCT images into anatomical compartments such as cortical and trabecular regions. However, motion artifacts can compromise cortical bone architecture and blur trabecular microstructures. In this study, we employed autocontour from the ``ORMIR\_ XCT" library \cite{Kuczynski2024}, a segmentation algorithm tailored to mimic the Image Processing Language (IPL) used by the HR-pQCT scanner, across all anatomical sites. To evaluate the impact of motion correction on segmentation performance, we compared standard image similarity metrics\cite{Taha2015}—Dice coefficient, Jaccard index, and Hausdorff distance—between motion-corrected and ground-truth images.The results are presented in Table \ref{table:5}. The Dice coefficient and Jaccard index quantify the similarity between the motion-corrected segmentation and the ground truth, with higher values indicating better agreement. In contrast, the Hausdorff distance measures the maximum boundary deviation between the predicted and reference segmentations, where lower values denote improved accuracy. A uniform, empirically selected threshold was applied across all images, which may have introduced variability in the metric values across different models. Fig. \ref{fig:10review} presents the segmentation results obtained using autocontour on motion-corrected images produced by the four models evaluated in section \ref{ablation}. A progressive refinement of the cortical boundary is observed with increasing quality of motion correction. However, due to inherent limitations of the autocontour algorithm, a comprehensive investigation of HR-pQCT segmentation performance—both before and after motion correction—is required prior to drawing definitive conclusions.}

\begin{figure*}[t]

    \includegraphics[width=0.99\linewidth]{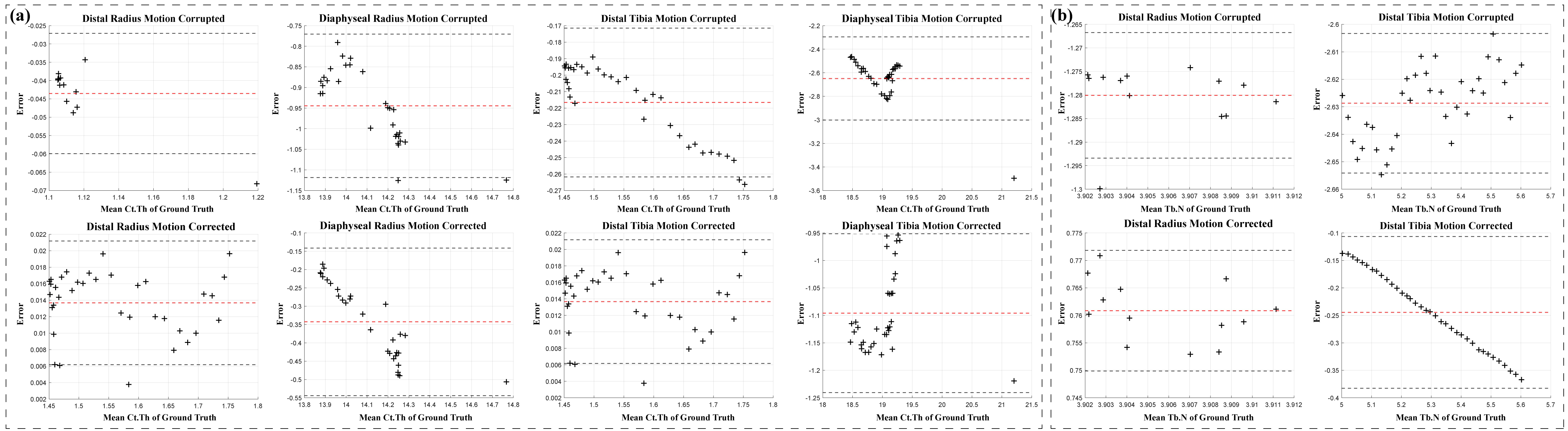}
    \caption{\textcolor{black}{Bland-Altman plots evaluating the agreement between ground truth and ESWGAN-GP–predicted bone geometry parameters relative to their mean ground truth values. (a) and (b) illustrate the effects of motion artifacts and their correction by ESWGAN-GP on cortical thickness (Ct.Th) and trabecular number (Tb.N), respectively. The first row shows measurements from motion-corrupted data, while the second presents those after correction. \textcolor{black}{The red dashed line represents the mean difference, while the black dashed lines indicate the 95\% limits of agreement, reflecting the expected range of variation in the prediction errors.} }}
    \label{fig:17 review}
\end{figure*}

Following segmentation, in vivo quantitative assessments of bone geometry, including parameters such as cortical thickness (Ct.Th), trabecular number (Tb.N), and bone mineral density (BMD) can be derived from HR-pQCT imaging. These measurements have been demonstrated to capture variations associated with age, sex, disease progression, and the effects of pharmaceutical interventions \cite{Whittier2020}. Accurate calculation of these quantitative parameters from HR-pQCT scans necessitates precise segmentation. To ensure automation and consistency across slices, autocontour and a fixed threshold were applied during the morphological operations throughout the segmentation process. This experiment was conducted on the source dataset, where ground truth measurements were available. Fig. \ref{fig:15review} demonstrates the performance of ESWGAN-GP in recovering Ct.Th in all of the four peripheral sites and Tb.N in Distal Radius and Tibia. Strong correlations were observed due to motion correction, with the exception of the Ct.Th measurement at the \textcolor{black}{Diaphyseal} Radius, where the lower correlation is attributed to an outlier resulting from suboptimal segmentation. Given the sensitivity of the correlation coefficient to outliers, we additionally present Bland-Altman plots in Fig. \ref{fig:17 review}. These plots display the differences between the ground truth and the motion-corrupted measurements (first row), as well as the differences between the ground truth and motion-corrected measurements (second row), plotted against the mean of the corresponding ground truth values. \textcolor{black}{A noticeable reduction in error is observed following motion correction. For instance, in the \textcolor{black}{Diaphyseal} Radius, the average absolute error in Ct.Th decreased from approximately 0.95 in the motion-corrupted data to 0.35 in the motion-corrected data. Similarly, in the \textcolor{black}{Diaphyseal} Tibia, the average error in Ct.Th was reduced from around 2.7 to 1.1. Comparable trends are evident in the distal sites for both Ct.Th and Tb.N measurements. An important observation from the Bland-Altman plots in Fig. \ref{fig:17 review} is that, although measurement errors are reduced, they do not cross zero, indicating that a complete elimination of motion-induced consequences is unlikely.}

\begin{figure}[h]

    \includegraphics[width=0.99\linewidth]{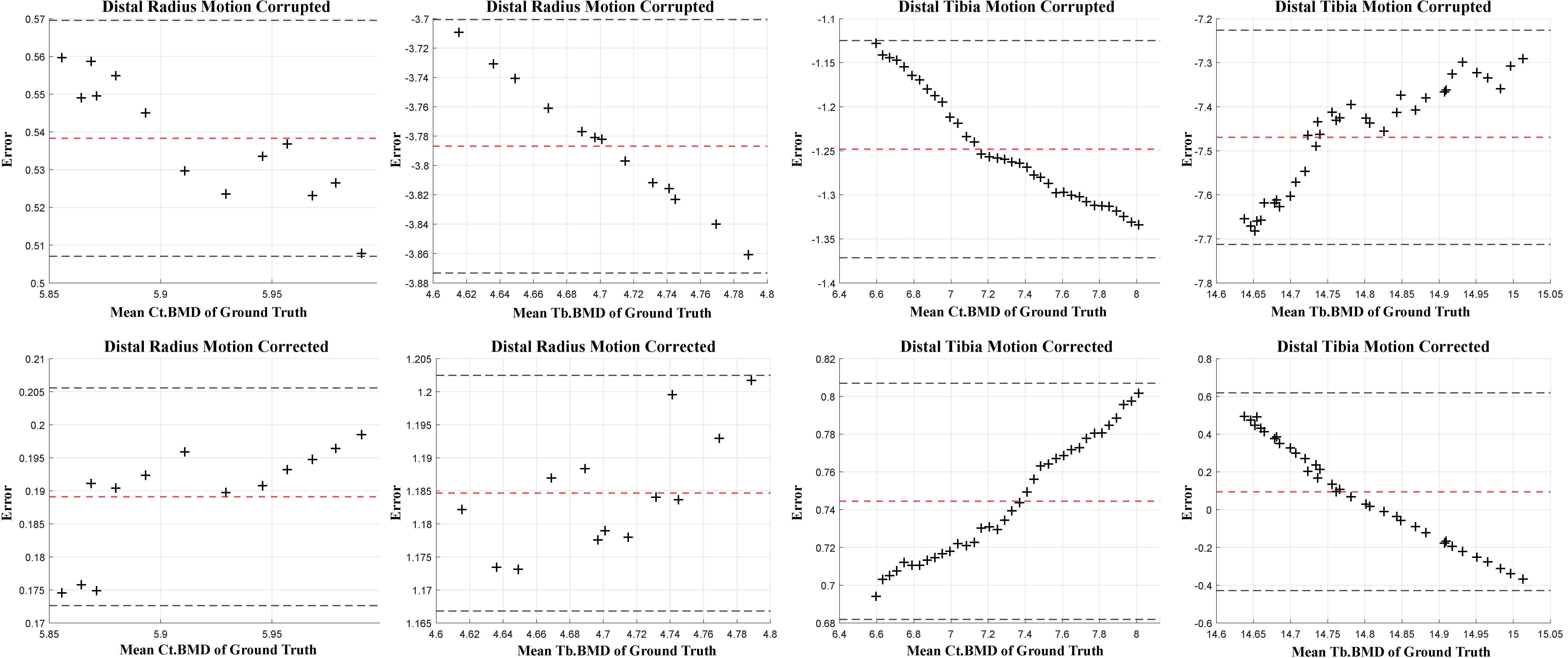}
    \caption{\textcolor{black}{Bland–Altman plots depicting the agreement between ground truth and ESWGAN-GP–predicted values for cortical BMD (Ct.BMD) and trabecular BMD (Tb.BMD). The plots display the prediction error as a function of the mean ground truth BMD. The first row shows measurements from motion-corrupted data, while the second presents those after correction. \textcolor{black}{Similar to Fig. \ref{fig:17 review}, the red dashed line represents the mean difference.}}}
    \label{fig:16 review}
\end{figure}

We additionally computed the mean bone mineral density (BMD) for both cortical (Ct.BMD) and trabecular (Tb.BMD) regions. This was achieved by applying the respective cortical and trabecular masks to the image data, followed by averaging the resulting values across the entire volume. Fig. \ref{fig:16 review} presents Bland–Altman plots illustrating the prediction errors of Ct.BMD and Tb.BMD relative to the ground truth BMD measurements. The analysis is conducted across both simulated corrupted and ESWGAN-GP-corrected volumes. A reduction in prediction error is observed in the motion-corrected scenarios. \textcolor{black}{For example, the mean Ct.BMD in the motion-corrupted Distal Radius data is 0.54, whereas it is reduced to 0.19 following motion correction, indicating a mitigation of motion-induced artifacts. In Tb.BMD of Distal Tibia, a few volumes have nearly crossed the zero line, indicating a substantial attenuation of motion artifacts. These findings collectively suggest that while motion correction has a beneficial impact on quantitative measurements, its capacity to fully resolve motion artifacts remains uncertain and warrants further investigation through more comprehensive studies.} 

Notably, \textcolor{black}{in Fig. \ref{fig:16 review},} the error appears to vary systematically with the mean BMD, demonstrating a tendency to underestimate BMD at lower mean values and overestimate it at higher ones. This pattern may be attributable to the automated segmentation process, which tends to under-segment (i.e., thin) the cortical and trabecular compartments in regions of lower BMD and over-segment (i.e., thicken) them in regions of higher BMD. Further investigations are necessary to obtain more accurate estimations of the quantitative measurements and, consequently, to enable a comprehensive evaluation of the motion correction performance.

\textcolor{black}{Despite its limitations, the proposed approach establishes the foundation for deep learning–based motion correction in HR-pQCT bone imaging and highlights its potential impact on downstream tasks, including image segmentation and quantitative parameter estimation.}
\section{Conclusion}

In conclusion, this study \textcolor{black}{optimizes} a sinogram-based approach to simulate in-plane rotational motion artifacts in HR-pQCT, to train an Edge-enhanced Self-attention Wasserstein Generative Adversarial Network with Gradient Penalty (ESWGAN-GP) that can effectively address motion artifacts in both simulated and real-world data. The integration of edge-enhancing skip connections, self-attention mechanisms, and a VGG-based perceptual loss contributes to the accurate reconstruction of fine microstructural features. This work lays the groundwork for the application of deep learning in motion correction for HR-pQCT, \textcolor{black}{potentially reducing patient rescans by up to 10\% per week}, a major challenge for the broader adoption of this modality. The utility of the deep learning framework, particularly the edge enhancer block, can be extended to quantitative computed tomography \textcolor{black}{(QCT)} and micro-computed tomography (micro-CT). \textcolor{black}{The practical applicability of the proposed study is constrained by its focus on in-plane rotational motion within the simulation framework. Future work will aim to incorporate a wider range of motion artifacts, including in-plane translation, out-of-plane rotation, and z-axis translation into the simulation process, which will be subsequently utilized for training deep learning models.}

\section{Acknowledgement}

All authors declare that they have no known conflicts of interest in terms of competing financial interests or personal relationships that could have an influence or are relevant to the work reported in this paper. The authors would like to thank Peter K. Jalaie for data acquisition, organization, \textcolor{black} {and Anika Mathur for organizing the codes.}

\bibliographystyle{ieeetr}
\bibliography{sn-bibliography_new.bib} %IEEEabrv,

\end{document}